\documentclass{article}

\usepackage[main,final]{neurips_2026}
\usepackage[utf8]{inputenc}
\usepackage[T1]{fontenc}
\usepackage{graphicx}
\usepackage{float}
\usepackage[hyperfootnotes=false]{hyperref}
\usepackage{url}
\usepackage{booktabs}
\usepackage{multirow}
\usepackage{tabularx}
\usepackage{amsfonts}
\usepackage{amsmath}
\usepackage{nicefrac}
\usepackage{microtype}
\usepackage[table]{xcolor}
\usepackage{listings}
\usepackage{algorithm}
\usepackage{algpseudocode}

\newcommand{\consilium}{\textsc{Consilium}}
\newcommand{\learnmethod}{\textsc{Manana}}
\newcommand{\egpa}{error-grounded prompt learning}

\newcolumntype{Y}{>{\raggedright\arraybackslash}X}
\definecolor{promptbg}{RGB}{246, 249, 253}
\definecolor{promptborder}{RGB}{127, 163, 205}
\definecolor{oursgreen}{RGB}{242, 250, 244}
\definecolor{bestgreen}{RGB}{218, 239, 223}
\definecolor{catgray}{RGB}{230, 230, 230}
\newcommand{\best}[2]{\cellcolor{bestgreen}\ensuremath{\mathbf{#1}_{\pm#2}}}
\lstdefinestyle{promptstyle}{
  basicstyle=\ttfamily\small,
  backgroundcolor=\color{promptbg},
  breakautoindent=false,
  breakatwhitespace=false,
  breakindent=0pt,
  breaklines=true,
  columns=fullflexible,
  frame=single,
  framerule=0.35pt,
  framesep=5pt,
  keepspaces=true,
  rulecolor=\color{promptborder},
  showstringspaces=false,
  xleftmargin=0pt,
  xrightmargin=0pt,
  aboveskip=0.45\baselineskip,
  belowskip=0.9\baselineskip
}
\title{Teaching LLMs to Recommend and Defer in Underrepresented Epilepsy Care}

\author{%
\normalfont\small
\begin{tabular}{c}
\textbf{Shreyas Rajesh}\textsuperscript{1,*},
\textbf{Kartik Sharma}\textsuperscript{1,*},
\textbf{Tonmoy Monsoor}\textsuperscript{1},
\textbf{Mehmet Yigit Turali}\textsuperscript{1},\\
\textbf{Richard Idro}\textsuperscript{3},
\textbf{Juliana Kayaga}\textsuperscript{3},
\textbf{Robert Sebunya}\textsuperscript{4},
\textbf{Tracy Tushabe Namata}\textsuperscript{3},\\
\textbf{Jessica Nichole Pasqua}\textsuperscript{2},
\textbf{Vwani Roychowdhury}\textsuperscript{1},
\textbf{Rajarshi Mazumder}\textsuperscript{2}\\[0.45em]
\textsuperscript{1}Electrical and Computer Engineering, University of California, Los Angeles\\
\textsuperscript{2}Neurology, University of California, Los Angeles\\
\textsuperscript{3}Makerere University \quad
\textsuperscript{4}St. Francis Hospital Nsambya\\[0.25em]
\textsuperscript{*}Equal contribution.\\[0.15em]
\raisebox{-0.15ex}{\includegraphics[height=0.95em]{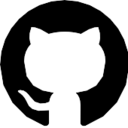}}\ \textsf{\bfseries GitHub}:
\href{https://github.com/roychowdhuryresearch/Manana}{\texttt{github.com/roychowdhuryresearch/Manana}}
\end{tabular}%
}

\hypersetup{
  pdftitle={Teaching LLMs to Recommend and Defer in Underrepresented Epilepsy Care},
  pdfauthor={Shreyas Rajesh, Kartik Sharma, Tonmoy Monsoor, Mehmet Yigit Turali, Richard Idro, Juliana Kayaga, Robert Sebunya, Tracy Tushabe Namata, Jessica Nichole Pasqua, Vwani Roychowdhury, Rajarshi Mazumder},
  pdfkeywords={clinical decision support, epilepsy, large language models, prompt learning, uncertainty, deferral},
}

\newcommand{\correspondence}{%
  \begin{tabular}{@{}l@{}}
  Correspondence: \href{mailto:kartiksharma@g.ucla.edu,shreyasrajesh38@g.ucla.edu}{\texttt{\{kartiksharma,shreyasrajesh38\}@g.ucla.edu}},\\
  \href{mailto:vwani@ee.ucla.edu}{\texttt{vwani@ee.ucla.edu}}, and
  \href{mailto:rmazumder@mednet.ucla.edu}{\texttt{rmazumder@mednet.ucla.edu}}.%
  \end{tabular}%
}

\begin{document}

\maketitle
\begingroup
\renewcommand{\thefootnote}{}
\footnotetext{\correspondence}
\endgroup

\begin{abstract}
Specialist epilepsy expertise is scarce in resource-constrained settings, creating a pressing need for decision-support tools that help frontline clinicians manage longitudinal treatment. Such systems must adapt to local prescribing practices and recognize when specialist review is needed. Most public medical AI benchmarks focus on high-income clinical settings, leaving prescribing practices, medication availability, and follow-up patterns in low-resource contexts largely unrepresented. We study this problem through a multidisciplinary collaboration focused on pediatric epilepsy care in Uganda. The task is to predict anti-seizure medication regimens from longitudinal unstructured notes collected by local clinicians across serial visits. General-purpose language models achieve non-trivial agreement with physician prescriptions, but neurologists' review of model reasoning traces shows that their errors reflect Western prescribing defaults that do not account for local clinical practices. We introduce \learnmethod{}, a \textbf{non-parametric prompt-learning framework} that learns how to reason about local prescribing decisions from a small patient-level training set. \learnmethod{} turns observed prescription errors into an auditable prompt memory, instantiated in single-agent and multi-agent variants, and outperforms classical ML models, direct LLM prompting, and prompt-optimization baselines across two independently collected Ugandan cohorts. The sequence of prompts produced during learning also provides a basis for estimating uncertainty. We exploit this through \textbf{Bayesian prompt averaging (BPA)}, which combines predictions across the learned prompt trajectory to produce prescription likelihoods and a deferral signal. Using the same \learnmethod{}-Multi trajectory, BPA improves top-1 physician-regimen agreement from 66\% to 70\% on Cohort~A and from 73\% to 75\% on Cohort~B while preserving top-3 coverage. More consequentially, it enables clinically meaningful selective prediction: the system can auto-handle the most confident half of cases at \textbf{95\% precision}, or the most confident quarter at \textbf{99\% precision}, while deferring lower-confidence cases for specialist review. These results suggest a path toward locally adapted clinical LLM systems that learn from limited site-specific data and reserve scarce specialist attention for the cases where uncertainty is highest.
\end{abstract}

\begin{figure*}[t]
\centering
\includegraphics[width=0.95\textwidth]{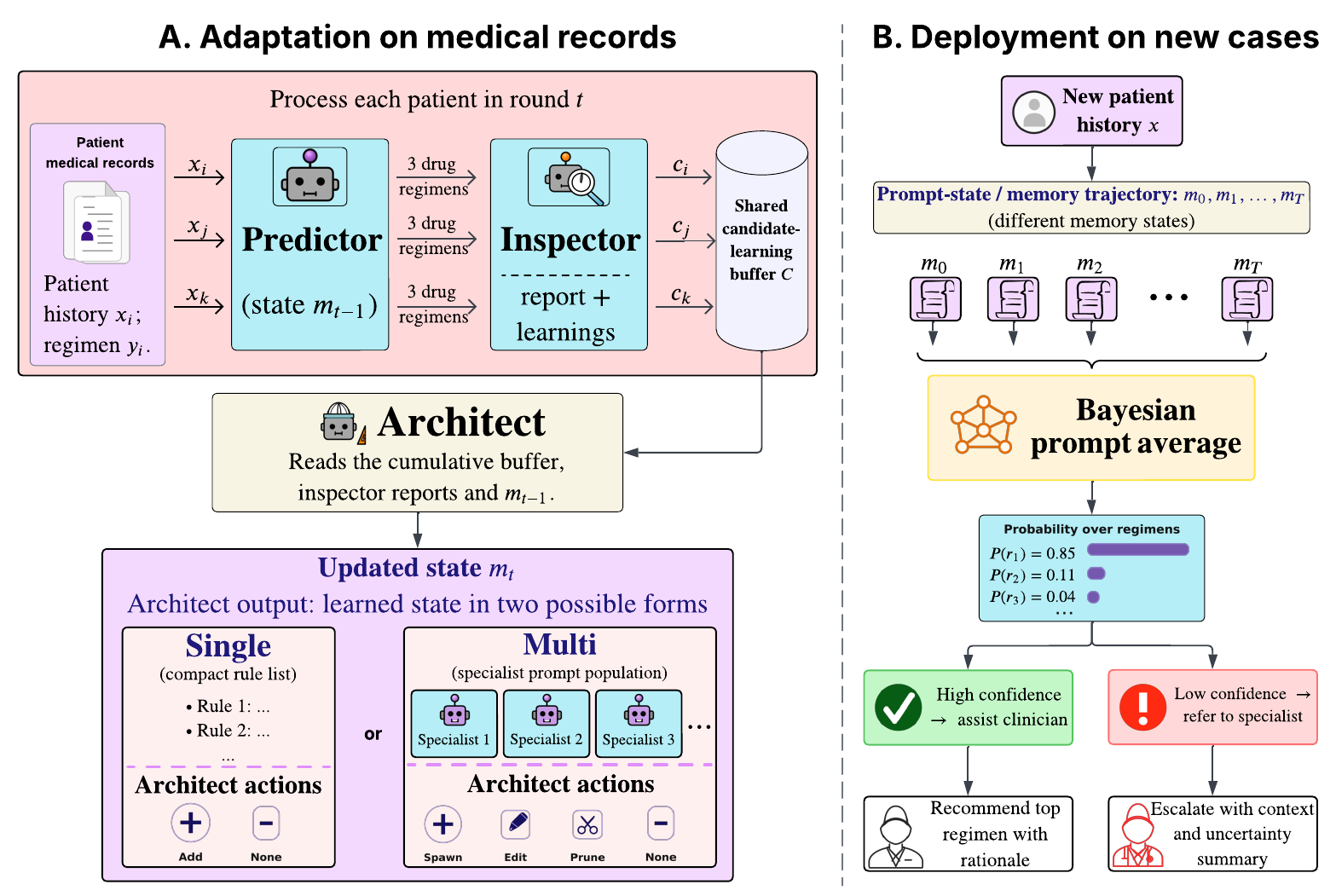}
\caption{\learnmethod{} workflow. (A) During adaptation, each batch of patient records is processed with the current memory state: the Predictor proposes three regimens, the Inspector compares them with the physician prescription revealed as supervision, and the proposed candidate learnings accumulate in an append-only buffer. Once per batch, the Architect consolidates recurring cross-case signals into the next shared memory, either as a Single correction-rule list or a Multi specialist-prompt population. (B) At inference, the learned memory trajectory is treated as an ensemble of prompt states. Bayesian prompt averaging converts their candidate regimen predictions into prescription probabilities, enabling high-confidence assistance or deferral to a specialist.}
\label{fig:manana_workflow}
\end{figure*}

\section{Introduction}

Neurological conditions are a major and unevenly distributed source of global health burden. More than three billion people live with neurological conditions worldwide, and over 80\% of neurological deaths and health loss occur in low- and middle-income countries (LMICs)~\citep{gbd2024nervoussystem}. Epilepsy affects around 0.8\% of the global population ($\sim$50 million people); nearly 80\% live in LMICs, and many do not receive the treatment they need~\citep{who2024epilepsy}. Specialist epilepsy care remains scarce in many LMICs, where diagnosis and treatment fall to generalist providers~\citep{who2017neurologyatlas}. Longitudinal epilepsy medication management in these settings is challenging because clinicians must adjust treatment over time despite incomplete diagnostic information, limited EEG or imaging access, medication stock-outs, comorbidities, adverse effects, and variable follow-ups. Frontline clinicians must decide when to initiate or switch anti-seizure medications, escalate doses, manage breakthrough seizures or side effects, and identify patients who need referral to specialists in an overburdened system. LLM-based decision support is therefore attractive: a system that relies on local clinical notes and assists with treatment decisions could extend specialist-informed reasoning to settings where such expertise is scarce~\citep{diessen2024epilepsyllm}.

Deployment, however, requires more than answering medical exam questions. Medical LLM agents have largely been evaluated on standardized QA, diagnosis, or EHR-operation benchmarks~\citep{kim2024mdagents,jiang2025medagentbench}. Moreover, medication recommendation systems typically use structured hospital data from high-income countries with high-resource settings such as MIMIC~\citep{shang2019gamenet,yang2021safedrug,yang2021micron,sun2022drugrec,fan2025flame,johnson2023mimiciv}. A realistic decision-support system in an underrepresented clinic needs two additional properties. First, its recommendations must be calibrated to local practice: drug availability, cost, and follow-up norms. Second, the system must surface uncertainty in medical decision-making, because scarce specialist attention should be directed toward cases where the model is least reliable.

We study these requirements in longitudinal epilepsy care in Uganda, where the task is to predict anti-seizure medication regimens across serial clinical visits. \textit{Here, we measure performance as agreement with the anti-seizure medication (ASM) regimen selected by the treating physician.} Each prediction is grounded in the clinic notes available for that patient, rather than structured diagnostic codes or medication histories from a high-resource EHR. Within this setting, we test whether an LLM can recover local prescribing practice from clinician notes and perform reliably across serial visits.

Direct prompting provides a useful baseline but not a deployable system: although a single-agent LLM achieves non-trivial agreement with physician prescriptions, neurologists' audit shows systematic errors driven by Western prescribing priors rather than Ugandan clinical practice. The problem is therefore not missing medical knowledge, but applying the wrong prior in the target health system. We ask whether an LLM, without parametric weight updates, can learn deployment-specific prescribing corrections from local cases and use the resulting adaptation trajectory to estimate uncertainty for deferral. This constraint matters because new clinical deployments need adaptation to local data without the retraining, validation, and audit burden of weight updates, especially where data and compute are limited.

% Direct prompting provides a useful baseline but not a deployable system. A single-agent LLM achieves non-trivial agreement with physician prescriptions, yet neurologist audit shows that its errors are systematic: it applies defaults consistent with Western prescribing priors rather than Ugandan clinic practice. The model has generic knowledge of anti-seizure medications, but it miscalibrates that knowledge to the deployment distribution. This distinction is central: the problem is not simply missing medical knowledge, but the failure to apply the right prior in the right health system.

% We therefore ask whether an LLM, without parametric weight updates, can acquire deployment-specific prescribing corrections from data alone, and whether the resulting adaptation process can also provide an uncertainty signal for deferral. This constraint matters because deployment across new regions requires adaptation to new local data; relying on parametric weight updates can limit that adaptability through retraining, validation, and audit burdens, especially where local data and compute are limited.

We introduce \learnmethod{}, an \egpa{} framework designed to address these questions. \learnmethod{} learns from patient notes and physician prescriptions, without clinician-written rules or expert review of model reasoning traces. \learnmethod{} is organized as a multi-agent learning framework: one agent proposes medication regimens, a second agent analyzes errors against physician prescriptions, and a third agent consolidates recurring lessons into an interpretable prompt memory. This loop turns observed failures into deployment-specific reasoning guidance while retaining an auditable text representation that clinicians can review. We present two variants: \learnmethod{}-Single, which learns a single shared set of learnings, and \learnmethod{}-Multi, which instantiates specialist agents with learned prompts.

To convert prompt learning into uncertainty-aware decision support, we propose Bayesian prompt averaging (BPA): Bayesian model averaging over learned prompt trajectories. Each learning round produces a distinct prompt state, and these prompt states form an ensemble of estimators over medication regimens. Because each state reflects a different stage of evidence accumulation, the trajectory captures a range of plausible local prescribing rules learned from the calibration data. Weighting these estimators by their posterior support yields prescription probabilities rather than only ranked predictions, enabling selective prediction: high-confidence cases can proceed without additional specialist review, while low-confidence cases can be deferred to clinicians for further review. This concentrates scarce specialist attention on the cases where the system is least certain.
We summarize our contributions as follows:
\vspace{-0.45em}
\begin{enumerate}
\setlength{\itemsep}{0pt}
\setlength{\topsep}{0pt}
\setlength{\parsep}{0pt}
\setlength{\parskip}{0pt}
\setlength{\partopsep}{0pt}
  \item We present \textbf{\learnmethod{}}, a non-parametric prompt-learning framework that learns how to reason about local prescribing decisions from patient clinical notes and physician prescriptions, instantiated in single-agent and multi-agent variants.
  \item We show that, using only a small patient-level training set, \learnmethod{} achieves strong held-out performance across two independently collected Ugandan epilepsy cohorts, outperforming classical ML models, a single-agent LLM, and existing prompt optimization methods.
  \item We further propose \textbf{Bayesian prompt averaging (BPA)}, a Bayesian model averaging framework over learned prompt trajectories that produces prescription probabilities and deferral signals, allowing high-confidence cases to proceed without additional specialist review while routing low-confidence cases for further clinical review.
\end{enumerate}
\vspace{-0.45em}

\section{\learnmethod{}}
\label{sec:learning}

\learnmethod{} adapts an LLM without weight updates by learning an explicit memory state that conditions future predictions. During the learning phase, the system observes a small set of local training cases, each pairing the patient history available before prescription with the regimen prescribed by the treating clinician. In our experiments, this learning set contains 50 patients; Section~\ref{sec:experiments} gives the full split and evaluation protocol.

\learnmethod{} is a multi-agent learning system with three components. The \emph{Predictor} is the regimen-producing LLM: given a patient history and the current memory state, it returns three candidate regimens over the 10-drug clinic action space. The \emph{Inspector} compares those regimens with the clinician prescription and writes structured reports and candidate learnings. The \emph{Architect} aggregates evidence across candidate learnings and recent reports, then updates the memory state through the variant-specific update rule. The goal is to learn a sequence of memory states $m_0,m_1,\ldots,m_T$ that capture recurring deployment-specific prescribing corrections from these examples. All three components are LLM calls implemented with the same base model and role-specific prompts, so learning proceeds through memory-state updates rather than parameter updates. Figure~\ref{fig:manana_workflow} gives the end-to-end adaptation and inference workflow.

\subsection{Learning}

We represent each clinic visit as $(x_i,y_i)$, where $x_i$ is the cumulative patient history available before the prescribing decision and $y_i$ is the physician-prescription. Let $m_t$ denote the memory state after learning round $t$. The learning process starts with an empty memory, $m_0=\emptyset$. In round $t$, \learnmethod{} uses the current memory $m_{t-1}$ to process a batch of clinic visits $\mathcal{B}_t$. For each visit, Predictor returns
\[
\hat{y}_{i,1:3}=\mathrm{Pred}(x_i;m_{t-1}).
\]
The Inspector then compares those three candidates with the physician prescription and outputs a structured report $\rho_i$ and one \emph{candidate learning} $c_i$, a concise proposed lesson grounded in the note:
\[
(\rho_i,c_i)=\mathrm{Insp}(x_i,\hat{y}_{i,1:3},y_i).
\]
Candidate learnings are not immediately added to memory. They first enter an append-only \emph{evidence buffer} $\mathcal{C}$,
\[
\Delta\mathcal{C}_t=\bigl((i,c_i)\bigr)_{(x_i,y_i)\in\mathcal{B}_t},
\qquad
\mathcal{C}_t=\mathcal{C}_{t-1}\,\Vert\,\Delta\mathcal{C}_t,
\qquad
\mathcal{C}_0=\emptyset,
\]
where $i$ indexes the visit and $\Vert$ denotes append-only accumulation. The current batch reports are
\[
\mathcal{R}_t=\bigl((i,\rho_i)\bigr)_{(x_i,y_i)\in\mathcal{B}_t}.
\]
Unlike the evidence buffer, $\mathcal{R}_t$ is not persistent: it is recomputed for each batch and gives the Architect local diagnostic context for the current batch.
The Architect then updates memory as
\[
m_t=\mathrm{Arch}(m_{t-1},\mathcal{C}_t,\mathcal{R}_t),
\]
with the exact update rule determined by the memory variant. Section~\ref{sec:prompt_states} describes these update rules for the Single and Multi variants.

The separation between candidate learnings and memory updates is one of our main distinctions from existing prompt-optimization methods such as TextGrad~\citep{yuksekgonul2024textgrad}: the difference is not textual feedback, but the unit of update. TextGrad rewrites a mutable prompt variable via textual-gradient descent, producing global rules whose support is not tracked rule-by-rule across cases \learnmethod{} separates error diagnosis from memory update: the Inspector writes candidate learnings, the evidence buffer preserves them with case provenance, and the Architect applies a constrained memory update defined by the chosen variant. We return to the empirical implications of this design choice in Section~\ref{sec:results}.

\subsection{Memory Updates}
\label{sec:prompt_states}

\learnmethod{} uses the same Predictor--Inspector--Architect loop with two variants of Architect update.

\textbf{Single.}
In \learnmethod{}-Single, the memory state is a list of $L_t$ learned rules, $m_t=(r_1,\ldots,r_{L_t})$. The Architect either appends one new rule synthesized from recurring candidate learnings in the evidence buffer or leaves it unchanged:
\[
m_t \in \{m_{t-1},\,m_{t-1}\Vert r_{\mathrm{new}}\},
\qquad |m_t|-|m_{t-1}|\le1.
\]
Since the memory starts empty, this implies $L_t\le t$. The append action is allowed only when the proposed rule is supported by at least $N=2$ learnings originating from distinct clinic visits.

\textbf{Multi.}
In \learnmethod{}-Multi, the memory state is a set of active specialist agents,
\[
m_t=\{s_{\ell,t}:\ell\in\mathcal{A}_t\}.
\]
Here $\mathcal{A}_t$ is the set of active specialist agents after learning round $t$, and $s_{\ell,t}$ is the instruction for agent $\ell$ at that round. Each specialist agent is responsible for extracting one clinical signal from the patient history. For a case $x_i$, the active specialist agents produce observations
\[
o_{i,\ell,t}=s_{\ell,t}(x_i),
\qquad \ell\in\mathcal{A}_t.
\]
The Predictor receives these observations as case-specific context and produces the regimen predictions. The observations are not persistent memory; they are discarded after the case. At each learning round, it may perform up to two actions chosen from \textsc{spawn}, \textsc{edit}, \textsc{prune}, and \textsc{None}. Here \textsc{spawn} adds a new specialist agent, \textsc{edit} rewrites an existing specialist instruction, and \textsc{prune} removes a specialist agent. Spawn and edit actions require recurring support from distinct clinic visits, using candidate learnings in the evidence buffer together with the current batch reports. Prune actions are used when current reports indicate that a specialist is redundant, misleading, or violating its observation-only role. If the specialist set is empty, the Architect may perform one bootstrap \textsc{spawn} after the first batch so the Multi memory can begin. In our experiments, we prompt the Architect with a target budget $\mathcal{A}_{\max}=5$ to keep the specialist context small and inspectable, including an instruction to prune before spawning when the active set reaches this budget. Full prompt templates are provided with the released code; representative learned artifacts are summarized in Appendix~\ref{app:learned_artifacts}.

Both variants produce a memory trajectory rather than one final prompt. Different memory states can make different correct predictions on different patients. The next subsection uses this trajectory as the basis for uncertainty estimation.

\subsection{Bayesian Prompt Averaging}
\label{sec:bayes_method}

Clinical decision support requires a confidence score for each prediction, especially when specialist resources are sparse. Bayesian model averaging (BMA) provides this by averaging predictions over candidate models according to posterior support~\citep{mackay1992bayesian,hoeting1999bma}. For candidate models $m_1,\ldots,m_T$ and data $\mathcal{D}$,
\[
p(y\mid x,\mathcal{D})
=
\sum_{t=1}^{T}p(y\mid x,m_t)\,p(m_t\mid\mathcal{D}),
\qquad
p(m_t\mid\mathcal{D})
=
\frac{p(\mathcal{D}\mid m_t)p(m_t)}
{\sum_{q=1}^{T}p(\mathcal{D}\mid m_q)p(m_q)}.
\]

BPA instantiates the candidate models as learned memory states. The learning split $\mathcal{D}_{\mathrm{train}}$ produces a trajectory $m_1,\ldots,m_T$, where each state conditions the same Predictor and induces a different regimen predictor. We use the held-out validation set $\mathcal{D}_{\mathrm{val}}$ to estimate posterior support over this finite learned trajectory, following validation-based predictive averaging and stacking approaches that weight fixed predictors by held-out or out-of-sample predictive performance~\citep{wolpert1992stacked,yao2018stacking}. We retain the top $K\leq T$ states by validation marginal likelihood; setting $K=T$ recovers the full trajectory. After reindexing the retained states as $m_1,\ldots,m_K$, we write
\[
\ell_k=\log p(\mathcal{D}_{\mathrm{val}}\mid m_k),
\qquad
w_k =
\frac{\exp\{\ell_k/\tau\}}
{\sum_{q=1}^{K}\exp\{\ell_q/\tau\}},
\]
which is a temperature-smoothed normalized posterior weight under a uniform prior over retained states. BPA then computes
\[
p_{\mathrm{BPA}}(y\mid x)
=
\sum_{k=1}^{K} w_k\,p(y\mid x,m_k).
\]
It remains to define the memory-state predictive likelihood $p(y\mid x,m_k)$ and the validation marginal likelihood $p(\mathcal{D}_{\mathrm{val}}\mid m_k)$. For a new patient history $x$, the Predictor under memory state $m_k$ returns three candidate regimens $\hat{y}_{k,1:3}$. Since the LLM call does not return calibrated probabilities, we represent $p(y\mid x,m_k)$ with an empirical-Bayes-style candidate-position prior estimated from a small subset of the training set, $\pi=(0.85,0.11,0.04)$~\citep{robbins1956empirical}:
\[
p(y\mid x,m_k)=\sum_{j=1}^{3}\pi_j\mathbf{1}[\hat{y}_{k,j}=y].
\]

For $p(\mathcal{D}_{\mathrm{val}}\mid m_k)$, we run each retained memory state on every validation case and count position-1 hits $c_{k,1}$, position-2-or-3 hits $c_{k,>1}$, and misses $u_k$, with $h_k=c_{k,1}+c_{k,>1}$ and $\pi_{>1}=\pi_2+\pi_3$. With a $\mathrm{Beta}(1,1)$ prior on the top-3 hit probability, the integrated validation likelihood is
\[
p(\mathcal{D}_{\mathrm{val}}\mid m_k)
=
\pi_1^{c_{k,1}}\pi_{>1}^{c_{k,>1}}B(h_k+1,u_k+1).
\]
Appendix~\ref{app:bpa_ablations} gives the derivation and reports alternatives that relax the shared candidate-position prior. Substituting the candidate-position predictive likelihood gives
\[
p_{\mathrm{BPA}}(y \mid x)
=
\sum_{k=1}^{K} w_k \sum_{j=1}^{3}\pi_j\mathbf{1}[\hat{y}_{k,j}=y].
\]
For top-1 evaluation, we report the regimen with maximum posterior predictive mass; for top-3 evaluation, we report the three regimens with highest posterior predictive mass. For selective prediction, the posterior mass assigned to a reported regimen is its confidence score, allowing low-confidence cases to be routed for specialist review.

\section{Experimental Setup}
\label{sec:experiments}

\textbf{Clinical cohorts and preprocessing.}
We study two independently collected pediatric epilepsy cohorts from Ugandan referral centers staffed by specialist pediatric neurologists: Cohort~A contains 332 patients and 1{,}040 visits, and Cohort~B contains 367 patients and 1{,}509 visits. The records are clinician-authored narrative outpatient notes from longitudinal epilepsy care, not structured high-resource EHR records such as MIMIC-IV~\citep{johnson2023mimiciv}. They document serial medication-management trajectories, including continuation, stopping, switching, adjunctive therapy, and escalation to polytherapy, rather than isolated one-time drug choices. Because regimen complexity varies across visits, the main analyses separate \textbf{monotherapy} visits, where the physician-selected active regimen contains one ASM, from \textbf{polytherapy} visits, where it contains two or more ASMs. The cohorts share a low-resource clinical context but differ in documentation format, clinicians, patient characteristics, and prescribing patterns, supporting a transfer test across related-but-shifted environments. LLM-assisted preprocessing separated each visit into pre-prescription clinical input and prescribed-regimen target under a no-leakage rule, with manual audit; Appendix~\ref{app:cohort_characteristics} gives preprocessing details, cohort characteristics, regimen-complexity distributions, and the 10-drug clinic action space. Main experiments use visits 1--3, where follow-up remains sufficiently populated for stable cohort-level estimates.

\textbf{Prediction task and metrics.}
For each visit, each method receives the current pre-prescription clinic note, prior visit notes, and prior prescribed regimens, then returns three possible anti-seizure medication (ASM) regimens. An ASM regimen is the active post-visit set of anti-seizure medications; evaluation uses only normalized active drug sets. Thus, performance is measured as agreement with the ASM regimen selected by the treating physician on the ground. The primary metric is exact match@3 (EM@3): a visit is correct if any proposed ASM regimen exactly matches the physician-prescribed regimen after medication-name normalization. For ranked outputs, Top-1 evaluates only the highest-ranked regimen, while Top-3/EM@3 evaluates whether any of the three returned regimens matches the physician-prescribed regimen. We also report maximum Jaccard similarity as partial-credit set overlap, and stratify by physician-prescribed monotherapy versus polytherapy because aggregate scores can hide different behavior on single-drug and combination regimens.

\textbf{Baselines and comparators.}
We compare \learnmethod{} against standard prompting baselines, classical predictors trained on extracted clinical features, the EpiPick epilepsy rule-based comparator for monotherapy, and prompt-optimization baselines including TextGrad, ExpeL, and DSPy~\citep{asadi2020pragmatic,epipick,yuksekgonul2024textgrad,zhao2024expel,khattab2023dspy}; we optimize the DSPy program with the GEPA optimizer~\citep{agrawal2025gepa}. Full baseline definitions and implementation details are in Appendix~\ref{app:main_results}.

\textbf{Implementation details.}
All learning methods train and validate on Cohort~A, with Cohort~B held out for transfer evaluation. We use a fixed 70-patient Cohort~A learning pool; each of five seeds stratifies it into 50 training and 20 validation patients, with patient-level separation. Training processes 150 visits in 15 update rounds, and reported seeded results are mean $\pm$ standard deviation over seed-level test scores. Unless otherwise stated, experiments use OpenAI gpt-oss-120b~\citep{openai2025gptoss} as the base LLM. Model inference ran on four NVIDIA RTX A6000 GPUs, with AWS Bedrock used for additional inference needs. For BPA, validation scoring retains $K=5$ memory states and uses $\tau=5$; duplicate regimens are merged by posterior predictive mass. All LLM-based methods use the same output schema and regimen parser. Full prompts are provided with the released code; additional model experiments are in Appendices~\ref{app:llm_backbone_sweep} and \ref{app:cross_model_transfer}.

\section{Results and Discussion}
\label{sec:results}

We evaluate \learnmethod{} on the longitudinal ASM regimen prediction task defined in Section~\ref{sec:experiments}: methods learn from Cohort~A training patients and are tested on held-out Cohort~A patients and the independently collected Cohort~B. \textbf{Physician-regimen agreement (EM@3)} (Table~\ref{tab:learning_main}): \learnmethod{} improves over the Base Prompt, the doctor-engineered Single-agent prompt, and classical baselines across most visits and regimen strata, while remaining competitive with or stronger than prompt-optimization methods. The dominant difficulty is regimen complexity: monotherapy is consistently easier than polytherapy because exact-match evaluation for a single drug only requires recovering one active medication, whereas polytherapy requires recovering the full drug combination. Appendix~\ref{app:main_results_interpretation} adds three interpretation checks: paired confidence intervals quantify uncertainty in method comparisons, a previous-regimen copy baseline tests whether high EM@3 mainly reflects unchanged prescriptions, and an improvement-stratified subset asks whether physician-regimen agreement remains strong on visits associated with documented seizure reduction or seizure freedom. \textbf{Regularization effect of \learnmethod{}}: the empirical gap between \learnmethod{} and prompt-rewrite baselines is consistent with a difference in the learned artifact, not only a difference in the underlying LLM. Candidate lessons must recur across patients before the Architect commits them, so \learnmethod{}-Single learns compact evidence-gated prescribing rules, while \learnmethod{}-Multi can spawn bounded specialist agents that surface recurring clinical signals. Appendix~\ref{app:learned_artifacts} shows these artifacts side by side with TextGrad and ExpeL: TextGrad learns detailed but weakly grounded global rules, ExpeL learns useful but more generic memories, and \learnmethod{} learns auditable rules or specialist signal extractors.

\begin{table*}[t]
\centering
\small\textbf{}
\setlength{\tabcolsep}{5pt}
\renewcommand{\arraystretch}{0.98}
\caption{Main EM@3 results by cohort and regimen complexity. Values are top-3 exact match percentages; prompt-optimization rows report mean $\pm$ std over independent seeds where applicable. Light green rows mark \learnmethod{} variants; darker cells mark the best value in each column.}
\label{tab:learning_main}
\begin{tabularx}{\textwidth}{@{}l Y c c c c c c@{}}
\toprule
& & \multicolumn{3}{c}{Monotherapy} & \multicolumn{3}{c}{Polytherapy} \\
\cmidrule(lr){3-5} \cmidrule(lr){6-8}
Cohort & Method & V1 & V2 & V3 & V1 & V2 & V3 \\
\midrule
\rowcolor{catgray}
\multirow{14}{*}{A}
  & \multicolumn{7}{c}{\textit{Expert System}} \\
  & EpiPick                  & 52.9 & 49.1 & 51.3 & --- & --- & --- \\
\addlinespace[0.15em]
  \rowcolor{catgray}
  & \multicolumn{7}{c}{\textit{Classical methods}} \\
  & Naive Bayes            & $71.3_{\pm2.1}$ & $67.2_{\pm3.7}$ & $74.9_{\pm2.9}$ & \best{35.8}{1.1} & $40.0_{\pm0.9}$ & $37.7_{\pm2.3}$ \\
  & XGBoost                  & $70.2_{\pm1.0}$ & $64.1_{\pm1.1}$ & $71.1_{\pm0.3}$ & $33.6_{\pm2.7}$ & $40.3_{\pm1.6}$ & $43.1_{\pm2.0}$ \\
\addlinespace[0.15em]
  \rowcolor{catgray}
  & \multicolumn{7}{c}{\textit{Direct LLM prompting}} \\
  & Base Prompt (0-learning) & $44.4_{\pm1.8}$ & $56.4_{\pm1.3}$ & $59.3_{\pm0.3}$ & $10.1_{\pm1.9}$ & $29.3_{\pm3.3}$ & $30.2_{\pm1.3}$ \\
  & Single-agent             & $78.4_{\pm1.1}$ & $84.5_{\pm1.9}$ & $90.3_{\pm2.4}$ & $14.5_{\pm1.2}$ & $40.2_{\pm0.7}$ & $42.3_{\pm2.5}$ \\
\addlinespace[0.15em]
  \rowcolor{catgray}
  & \multicolumn{7}{c}{\textit{Prompt optimization}} \\
  & DSPy (GEPA optimizer)                     & $87.1_{\pm1.8}$ & $91.3_{\pm4.1}$ & $95.0_{\pm3.2}$ & $24.5_{\pm0.9}$ & $58.7_{\pm2.9}$ & $70.2_{\pm3.4}$ \\
  & ExpeL                    & $67.2_{\pm2.4}$ & $81.0_{\pm3.9}$ & $82.7_{\pm3.7}$ & $18.9_{\pm1.1}$ & $28.3_{\pm1.4}$ & $42.9_{\pm2.5}$ \\
  & TextGrad                 & $87.4_{\pm2.3}$ & $88.1_{\pm5.2}$ & $87.3_{\pm4.4}$ & $21.9_{\pm4.1}$ & $50.7_{\pm4.3}$ & $57.6_{\pm3.0}$ \\
\rowcolor{oursgreen}
  & \learnmethod{}-Single & $87.0_{\pm0.4}$ & $92.1_{\pm2.5}$ & $93.2_{\pm2.1}$ & $25.3_{\pm1.0}$ & $57.0_{\pm5.6}$ & $64.1_{\pm3.9}$ \\
\rowcolor{oursgreen}
  & \learnmethod{}-Multi & \best{89.7}{3.2} & \best{95.3}{2.8} & \best{97.0}{2.1} & $30.8_{\pm1.9}$ & \best{61.7}{8.4} & \best{72.6}{11.1} \\
\midrule
\rowcolor{catgray}
\multirow{14}{*}{B}
  & \multicolumn{7}{c}{\textit{Expert System}} \\
  & EpiPick                  & 58.0 & 61.4 & 58.1 & --- & --- & --- \\
\addlinespace[0.15em]
  \rowcolor{catgray}
  & \multicolumn{7}{c}{\textit{Classical methods}} \\
  & Naive Bayes            & $37.7_{\pm2.6}$ & $40.9_{\pm3.2}$ & $38.2_{\pm2.5}$ & $4.4_{\pm1.2}$ & $2.5_{\pm0.5}$ & $4.0_{\pm1.3}$ \\
  & XGBoost                  & $41.5_{\pm2.6}$ & $45.5_{\pm2.2}$ & $45.3_{\pm2.3}$ & $10.4_{\pm1.4}$ & $7.1_{\pm0.6}$ & $4.6_{\pm0.3}$ \\
\addlinespace[0.15em]
  \rowcolor{catgray}
  & \multicolumn{7}{c}{\textit{Direct LLM prompting}} \\
  & Base Prompt (0-learning) & $53.1_{\pm0.5}$ & $57.6_{\pm0.7}$ & $58.4_{\pm1.1}$ & $24.8_{\pm3.6}$ & $29.1_{\pm1.6}$ & $30.4_{\pm1.6}$ \\
  & Single-agent             & $76.7_{\pm0.6}$ & $82.6_{\pm2.3}$ & $83.8_{\pm2.4}$ & $39.4_{\pm2.4}$ & $49.4_{\pm1.4}$ & $54.0_{\pm1.1}$ \\
\addlinespace[0.15em]
  \rowcolor{catgray}
  & \multicolumn{7}{c}{\textit{Prompt optimization}} \\
  & DSPy (GEPA optimizer)                     & $81.8_{\pm4.2}$ & $93.3_{\pm2.1}$ & \best{94.8}{1.1} & $44.8_{\pm0.8}$ & $59.6_{\pm3.7}$ & $62.7_{\pm2.9}$ \\
  & ExpeL                    & $83.1_{\pm3.8}$ & $88.5_{\pm4.8}$ & $86.5_{\pm4.5}$ & $32.0_{\pm2.6}$ & $56.2_{\pm3.3}$ & $64.3_{\pm4.6}$ \\
  & TextGrad                 & $74.0_{\pm9.9}$ & $87.7_{\pm4.7}$ & $87.3_{\pm4.2}$ & $39.0_{\pm3.3}$ & $52.4_{\pm8.6}$ & $57.3_{\pm8.5}$ \\
\rowcolor{oursgreen}
  & \learnmethod{}-Single & $82.2_{\pm5.0}$ & $92.8_{\pm2.7}$ & $92.2_{\pm2.2}$ & \best{48.2}{1.7} & \best{65.2}{3.8} & $66.4_{\pm4.7}$ \\
\rowcolor{oursgreen}
  & \learnmethod{}-Multi & \best{87.3}{2.9} & \best{95.1}{1.6} & $93.2_{\pm1.9}$ & $40.2_{\pm4.6}$ & $61.1_{\pm5.8}$ & \best{69.8}{7.5} \\
\bottomrule
\end{tabularx}
\end{table*}

\textbf{Uncertainty and deferral} (Tables~\ref{tab:ensemble_accuracy} and~\ref{tab:selective}, Figure~\ref{fig:bpa_uncertainty}): We next evaluate whether the learned trajectory can support uncertainty-aware decision support rather than only higher EM@3. Table~\ref{tab:ensemble_accuracy} compares three ways of weighting the same \learnmethod{}-Multi prompt trajectory. The methods yield similar Top-1 and Top-3 agreement, indicating that the deferral signal is not dependent on a particular weighting rule. A simple majority vote on this trajectory attains 66\%/83\% Top-1/Top-3 on Cohort~A and 73\%/88\% on Cohort~B; it therefore matches the trajectory's Top-3 coverage but does not provide a prescription distribution for confidence-based deferral. This confidence signal becomes a practical deferral policy in Table~\ref{tab:selective}: a clinic can choose a deferral operating point based on available specialist capacity, trading coverage for physician-regimen agreement among non-deferred cases. For example, on Cohort~B, \learnmethod{}-Multi reaches 99\% top-1 agreement on the most confident 25\% of cases and 95\% agreement on the most confident 50\%. Figure~\ref{fig:bpa_uncertainty} shows the corresponding confidence calibration and deferral curve.

\begin{table}[ht]
\centering
\caption{BPA weighting ablation for \learnmethod{}-Multi on held-out test sets. Each row ensembles the same retained Multi prompt trajectory: Linear weights states by validation top-3 accuracy, Beta-Binomial BPA uses the main-text marginal-likelihood weighting, and Dirichlet BPA estimates state-specific candidate-position distributions. Gap is the difference between mean top-1 confidence on correct and incorrect predictions; P@25\% and P@50\% are top-1 precision among the most confident cases.}
\label{tab:ensemble_accuracy}
\small
\begin{tabular}{llccccc}
\toprule
Cohort & Weighting & Top-1 & Top-3 & Gap & P@25\% & P@50\% \\
\midrule
\multirow{3}{*}{A}
& Linear             & 70\% & \textbf{84\%} & 0.191 & 96\% & 90\% \\
& Beta-Binomial BPA  & 70\% & 83\% & \textbf{0.194} & 96\% & 90\% \\
& Dirichlet BPA      & 70\% & \textbf{84\%} & 0.187 & \textbf{97\%} & 90\% \\
\midrule
\multirow{3}{*}{B}
& Linear             & \textbf{76\%} & 88\% & 0.208 & 99\% & 95\% \\
& Beta-Binomial BPA  & 75\% & 88\% & \textbf{0.214} & 99\% & 95\% \\
& Dirichlet BPA      & 75\% & 88\% & 0.203 & 99\% & 95\% \\
\bottomrule
\end{tabular}
\end{table}

\begin{table}[ht]
\centering
\caption{Selective prediction from \learnmethod{}-Multi Beta-Binomial BPA confidence. Precision denotes top-1 exact-match accuracy within the retained subset. Higher confidence thresholds produce higher precision at lower coverage, giving a practical deferral signal.}
\label{tab:selective}
\small
\begin{tabular}{lcccc}
\toprule
& \multicolumn{2}{c}{Cohort A} & \multicolumn{2}{c}{Cohort B} \\
\cmidrule(lr){2-3} \cmidrule(lr){4-5}
Threshold / subset & Precision & Coverage & Precision & Coverage \\
\midrule
Confidence $\geq 0.95$ & 97\% & 14\% & 99\% & 18\% \\
Confidence $\geq 0.90$ & 94\% & 33\% & 96\% & 40\% \\
Confidence $\geq 0.85$ & 86\% & 56\% & 92\% & 56\% \\
\midrule
Top 25\% most confident & 96\% & 25\% & 99\% & 25\% \\
Top 50\% most confident & 90\% & 50\% & 95\% & 50\% \\
All cases               & 70\% & 100\% & 75\% & 100\% \\
\bottomrule
\end{tabular}
\end{table}

The confidence and deferral curves are reported in Appendix~\ref{app:bpa_confidence}.

\textbf{Ablations, expert-designed comparison, and clinician review.} We run extensive ablations to test whether the result depends on a single component or model choice: Appendix~\ref{app:component_ablations} ablates the \learnmethod{} loop components, Appendix~\ref{app:cross_model_transfer} tests transfer from the 120B model to the 20B model, Appendix~\ref{app:llm_backbone_sweep} repeats the learning setup across LLM backbones, Appendix~\ref{app:bpa_ablations} varies the BPA weighting model, and Appendix~\ref{app:mimic} instantiates the method on MIMIC-IV as a public-data reproducibility check. We also include \consilium{} as a strong expert-designed reference system: neurologist review of single-agent failures was used to define specialist clinical lenses, which were then implemented as a hand-built multi-agent comparator. Appendix~\ref{app:consilium_reference} describes the audit and specialist-agent design, Appendix~\ref{app:consilium_ablations} tests whether the council is reducible, and Table~\ref{tab:end_main_em3} compares \consilium{} with \learnmethod{} and BPA. Appendix~\ref{app:manana_clinician_review} adds a qualitative neurologist review of \learnmethod{}-Multi against \consilium{}, focused on clinical coherence rather than another aggregate accuracy number. Together, these comparisons show that specialist feedback can produce a strong hand-designed system, but \consilium{} does not learn from site-specific supervision and does not by itself yield posterior regimen probabilities or a confidence-based deferral policy.

% Together, these results support the central claim: deployment-specific clinical correction signals can be learned from data, represented as auditable prompt memory, and converted into uncertainty-aware recommendations.

\section{Related Work}
\label{sec:related}

\textbf{LLMs for clinical decision support.}
Clinical LLM systems have been evaluated on medical QA, simulated EHR workflows, medication prediction, epilepsy-variable extraction, and medication-attribute extraction across heterogeneous EHR systems~\citep{singhal2023medpalm,nori2023medprompt,kim2024mdagents,jiang2025medagentbench,kraljevic2024foresight,williams2024llmed,lee2025epilepsyextract,fei2025medication}, while ASM-response and medication-recommendation models typically use coded or structured EHR features~\citep{hakeem2022epilepsy,shang2019gamenet,yang2021safedrug,sun2022drugrec}. These systems do not learn site-specific prescribing rules from a small clinic dataset, recover complete serial ASM regimens from narrative outpatient notes, or decide when to defer to scarce specialist review.

\textbf{Prompt optimization and self-improving agents.}
Prompt search, textual-gradient methods, and reflection-style agents adapt LLM behavior through text artifacts built from feedback~\citep{zhou2023ape,yang2024opro,khattab2023dspy,yuksekgonul2024textgrad,shinn2023reflexion,zhao2024expel}; we compare directly against DSPy, TextGrad, and ExpeL. For clinical prescription learning, the failure mode is the unit of update: salient individual cases can become brittle rules, and self-critiques can reinforce wrong priors~\citep{sclar2024quantifying,huang2024selfcorrect}. \learnmethod{} instead separates error diagnosis from memory commitment, allowing the Architect to commit only recurring cross-patient patterns.

\textbf{Uncertainty and selective prediction.}
Prior work on calibration, Bayesian ensembling, conformal prediction, and learning-to-defer studies when models should abstain or hand off to humans~\citep{guo2017calibration,hoeting1999bma,lakshminarayanan2017deepensembles,shafer2008conformal,madras2018defer}. BPA uses a different source of uncertainty: the learned prompt trajectory itself, treating prompt states across rounds as local estimators whose posterior concentration gives a practical deferral signal. We do not claim prospective conformal validity; we test whether these trajectories can route uncertain prescriptions to specialist review. A more expansive version of the related work expanding on each of these areas is provided in Appendix~\ref{app:related_extended}.
\section{Limitations and Future Work}
\label{sec:limitations}

While our results show promise for using LLM systems to learn local prescribing patterns in low-resource clinical settings, several limitations of the current study guide future work. First, BPA gives a retrospective confidence score, not a deployment-ready deferral rule. The deferral thresholds in this paper are chosen on held-out data to show the coverage--accuracy tradeoff. In a clinic, the threshold would need to be chosen before use, checked over time as the patient mix changes, and tied to a clear review workflow for deferred cases. The confidence scores and learned memories produced here would help such a workflow by identifying low-confidence visits for specialist review and exposing the local prescribing rules the system is using.

Second, our endpoint is agreement with the ASM regimen selected by the treating physician, not proof that the regimen is clinically optimal. Local prescriptions reflect clinical judgment, but also formulary limits, availability, affordability, and follow-up constraints. A model that matches local practice may therefore also learn local resource constraints. Future work should include a thorough clinical analysis of the memories and specialist agents learned by \learnmethod{}, as well as the reasoning produced by the resulting models. A complementary direction, and part of our ongoing work, is to predict seizure freedom or seizure reduction directly from the longitudinal record, either alongside physician-regimen agreement or as an outcome-conditioned extension of the current framework.

\section*{Responsible-use statement}

This system is not validated for autonomous prescribing. Potential harms include unsafe medication continuation or switching, overconfidence under distribution shift, amplification of inequitable local resource constraints, privacy loss, and instruction-following failures from clinical text. Any prospective deployment would require de-identification and access controls, clinician oversight for every recommendation, site-specific prospective validation, pre-specified referral thresholds on separate calibration data, calibration and drift monitoring, audit logs, and mandatory specialist review for low-confidence or out-of-scope cases.

\section*{Acknowledgments and Disclosure of Funding}

This work was supported by a Global Health Seed Grant from the David Geffen School of Medicine at UCLA. Rajarshi Mazumder was supported by the Fogarty International Center of the National Institutes of Health under Award Number K01TW012178. The content is solely the responsibility of the authors and does not necessarily represent the official views of the National Institutes of Health.

The authors declare no competing interests.

\bibliographystyle{plainnat}
\bibliography{references}

\appendix

\newpage
\section*{Appendix}

Our appendix is structured as follows:
\begin{enumerate}
    \item Appendix~\ref{app:related_extended}: Extended Related Work.
    \item Appendix~\ref{app:cohort_characteristics}: Clinical Cohorts and Distribution Shift.
    \item Appendix~\ref{app:main_results}: Baseline Definitions.
    \item Appendix~\ref{app:main_results_interpretation}: Interpreting the Main-Results Evaluation.
    \item Appendix~\ref{app:consilium_reference}: Single-Agent Audit and Expert System \consilium{}.
    \item Appendix~\ref{app:cross_model_transfer}: Cross-Model Transfer Learning.
    \item Appendix~\ref{app:llm_backbone_sweep}: Additional Open Source Model Experiments.
    \item Appendix~\ref{app:component_ablations}: \learnmethod{} Component Ablations.
    \item Appendix~\ref{app:manana_clinician_review}: Clinician Review of \learnmethod{}.
    \item Appendix~\ref{app:bpa_ablations}: \learnmethod{} Bayesian Prompt Averaging Ablations.
    \item Appendix~\ref{app:mimic}: MIMIC-IV.
    \item Appendix~\ref{app:consilium_ablations}: \consilium{} Council Ablations.
    \item Appendix~\ref{app:learned_artifacts}: Learned Artifacts Across Optimization Methods.
\end{enumerate}
\clearpage

\section{Extended Related Work}
\label{app:related_extended}

This appendix expands the related-work discussion of Section~\ref{sec:related}, covering the broader literature in clinical LLM decision support, prompt optimization and self-improving agents, and uncertainty quantification.

\subsection{Clinical LLM Decision Support}
Large language models have shown strong performance on medical QA, biomedical reasoning, and patient-facing dialogue~\citep{singhal2023medpalm,nori2023medprompt}, and recent clinical-agent systems extend this work to multi-agent reasoning and simulated EHR workflows~\citep{kim2024mdagents,jiang2025medagentbench,chen2024rareagents}. These systems are usually evaluated on benchmark answers, diagnostic tasks, or simulated actions rather than longitudinal prescribing decisions in a specific clinic. More directly related work predicts medications from MIMIC clinical text~\citep{kraljevic2024foresight,johnson2023mimiciv}, evaluates LLM recommendations from emergency-department notes~\citep{williams2024llmed}, or extracts epilepsy variables from clinic letters~\citep{holgate2024llama,lee2025epilepsyextract}; recent reviews discuss the potential uses and risks of LLMs in epilepsy care~\citep{diessen2024epilepsyllm}. Parallel ASM-response and medication-recommendation models predict seizure freedom, single-drug response, or multi-drug sets from structured clinical features or coded EHRs~\citep{hakeem2022epilepsy,shang2019gamenet,yang2021safedrug,yang2021micron,sun2022drugrec,fan2025flame}. Recent global-health CDS work motivates evaluation in underrepresented settings~\citep{nature2025africaclinical,lancet2025lmicroadmap}, but leaves open the regime studied here: learning site-specific prescribing rules from a small clinic dataset, recovering complete serial ASM regimens from narrative outpatient notes, and identifying cases that should be deferred to scarce specialist review.

\subsection{Prompt Optimization and Self-Improving Agents}
There has been extensive work on adapting LLM behavior through text artifacts created from feedback, rather than through weight updates. Prompt-optimization methods search or rewrite prompts against task feedback, including APE~\citep{zhou2023ape}, OPRO~\citep{yang2024opro}, ProTeGi~\citep{pryzant2023protegi}, DSPy and MIPRO~\citep{khattab2023dspy,opsahlong2024mipro}, and TextGrad~\citep{yuksekgonul2024textgrad}. Agent-learning methods such as Reflexion and Self-Refine~\citep{shinn2023reflexion,madaan2023selfrefine}, Voyager~\citep{wang2023voyager}, and ExpeL~\citep{zhao2024expel} accumulate verbal lessons or skills from execution feedback. Related work explores structured workspaces for episodic memory (GSW; \citealp{rajesh2026gsw}), continual knowledge consolidation (Panini; \citealp{rajesh2026panini}), and research coordination with evidence verification (PANINI-DR; \citealp{turali2026paninidr}). Evolutionary program-search work studies a similar optimization pressure at the population level: HSEvo preserves candidate diversity when objective-driven search would otherwise collapse onto narrow solutions~\citep{dat2024hsevo}. These systems show that textual feedback can improve future LLM calls, and we compare directly against DSPy, TextGrad, and ExpeL. For clinical prescription learning, the central failure mode is the unit of update: prompt-search and reflection methods can turn salient individual cases into broad rules, and self-generated critiques may reinforce incorrect priors when the model is not externally grounded~\citep{sclar2024quantifying,huang2024selfcorrect}. \learnmethod{} instead separates error diagnosis from memory commitment: Inspectors write case-level failures, but the Architect commits only recurring cross-patient patterns, an evidence-gated update that targets the clinical risk that a single unusual patient becomes a brittle prescribing rule.

\subsection{Uncertainty Quantification and Selective Prediction}
Reliable clinical decision support requires a model to know when not to act. Neural models can be accurate while poorly calibrated, especially under dataset shift~\citep{guo2017calibration,ovadia2019trust}, and clinical prediction work treats calibration as a deployment requirement~\citep{vancalster2019calibration}. Standard uncertainty methods marginalize over plausible models, including Bayesian model averaging and deep ensembles~\citep{hoeting1999bma,lakshminarayanan2017deepensembles}, while recent LLM work studies confidence elicitation, self-evaluation, and prompt ensembles for black-box models~\citep{kadavath2022know,tian2023justask,xiong2024uncertainty,tonolini2024bayesian}. Selective prediction, reject-option classification, conformal prediction, risk control, and learning-to-defer formalize when a system should abstain or hand off to a human expert~\citep{chow1970reject,elyaniv2010selective,geifman2017selective,shafer2008conformal,bates2021risk,madras2018defer,mozannar2020defer,raghu2019second,kompa2021secondopinion}. BPA uses a different source of uncertainty: the learned prompt trajectory itself. It treats prompt states across learning rounds as an empirical distribution over local estimators, combines their prescription distributions, and uses posterior concentration as a practical deferral signal. We therefore do not claim prospective conformal validity; instead, we test whether self-learned prompt trajectories can route uncertain medication predictions to specialist review.

\section{Clinical Cohorts and Distribution Shift}
\label{app:cohort_characteristics}

\paragraph{Cohort description.}
This appendix provides source-format, preprocessing, and distribution-shift details for the two Ugandan cohorts introduced in Section~\ref{sec:experiments}. Cohort~A contains 332 patients and 1{,}040 visits and was provided as a clinic spreadsheet with demographics, notes, medication fields, and follow-up documentation distributed across columns. Cohort~B contains 367 patients and 1{,}509 visits and was provided as clinic PDF records.

\paragraph{Preprocessing.}
Records were de-identified before analysis. The raw records did not consistently separate clinical observations from treatment plans. We therefore used LLM-assisted preprocessing to split each visit into clinical context and prescription text. The splitter kept history, examination findings, seizure descriptions, current medications, treatment response, developmental information, and investigation results on the input side. It assigned prescriptions, dose changes, drug starts or stops, investigation orders, referrals, and follow-up plans to the output side. The exact preprocessing prompts are provided with the released code documentation.

\paragraph{Drug-label extraction.}
We normalized each prescription into structured labels over the 10-drug action space. A pharmacist-style extraction prompt mapped brand names, abbreviations, spelling variants, and local documentation variants to canonical drug names, and returned drugs prescribed and drugs stopped. A separate cleaning pass merged duplicated prescription information across free-text plans and structured medication fields while preserving the original clinical wording.

\paragraph{Leakage controls.}
All preprocessing prompts used an explicit no-leakage rule: treatment decisions from the current visit must not appear in the model input. Prior prescriptions are included only for visits before the prediction visit, because they are part of the longitudinal history available to the physician. We manually audited every preprocessed patient record against its source notes. Any ambiguous input--target splits were corrected before final evaluation. The resulting dataset was verified to preserve the intended task: predicting the regimen from information available before the prescribing decision.

\begin{table*}[htbp]
  \centering
  \small
  \setlength{\tabcolsep}{7pt}
  \renewcommand{\arraystretch}{1.08}
  \caption{Regimen complexity by cohort and visit for the main evaluation visits. ``None'' indicates that no medication from the 10-drug action space was extracted as prescribed at that visit.}
  \label{tab:cohort_regimen_complexity}
  \begin{tabular}{@{}lrrrrrr@{}}
  \toprule
  \textbf{Measure} & \multicolumn{3}{c}{Cohort A} & \multicolumn{3}{c}{Cohort B} \\
  \cmidrule(lr){2-4}\cmidrule(l){5-7}
  & V1 & V2 & V3 & V1 & V2 & V3 \\
  \midrule
  Num. visits & 332 & 332 & 332 & 367 & 366 & 343 \\
  Mono & 249 & 246 & 226 & 231 & 234 & 207 \\
  Poly & 67 & 79 & 98 & 97 & 128 & 126 \\
  None & 16 & 7 & 8 & 39 & 4 & 10 \\
  \bottomrule
  \end{tabular}
  \end{table*}

\begin{figure*}[htbp]
\centering
\includegraphics[width=0.82\textwidth]{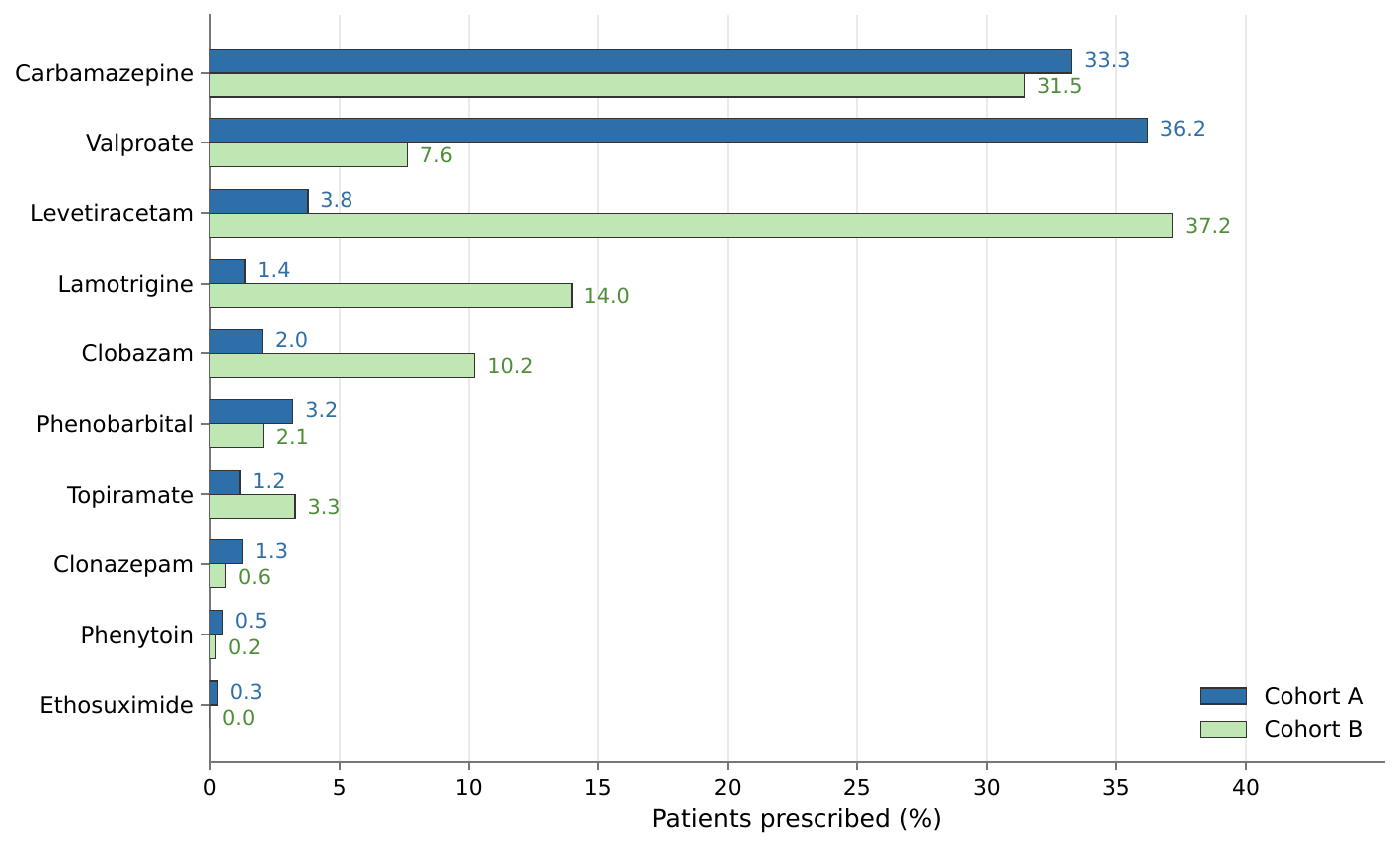}
\vspace{0.8em}
\includegraphics[width=0.82\textwidth]{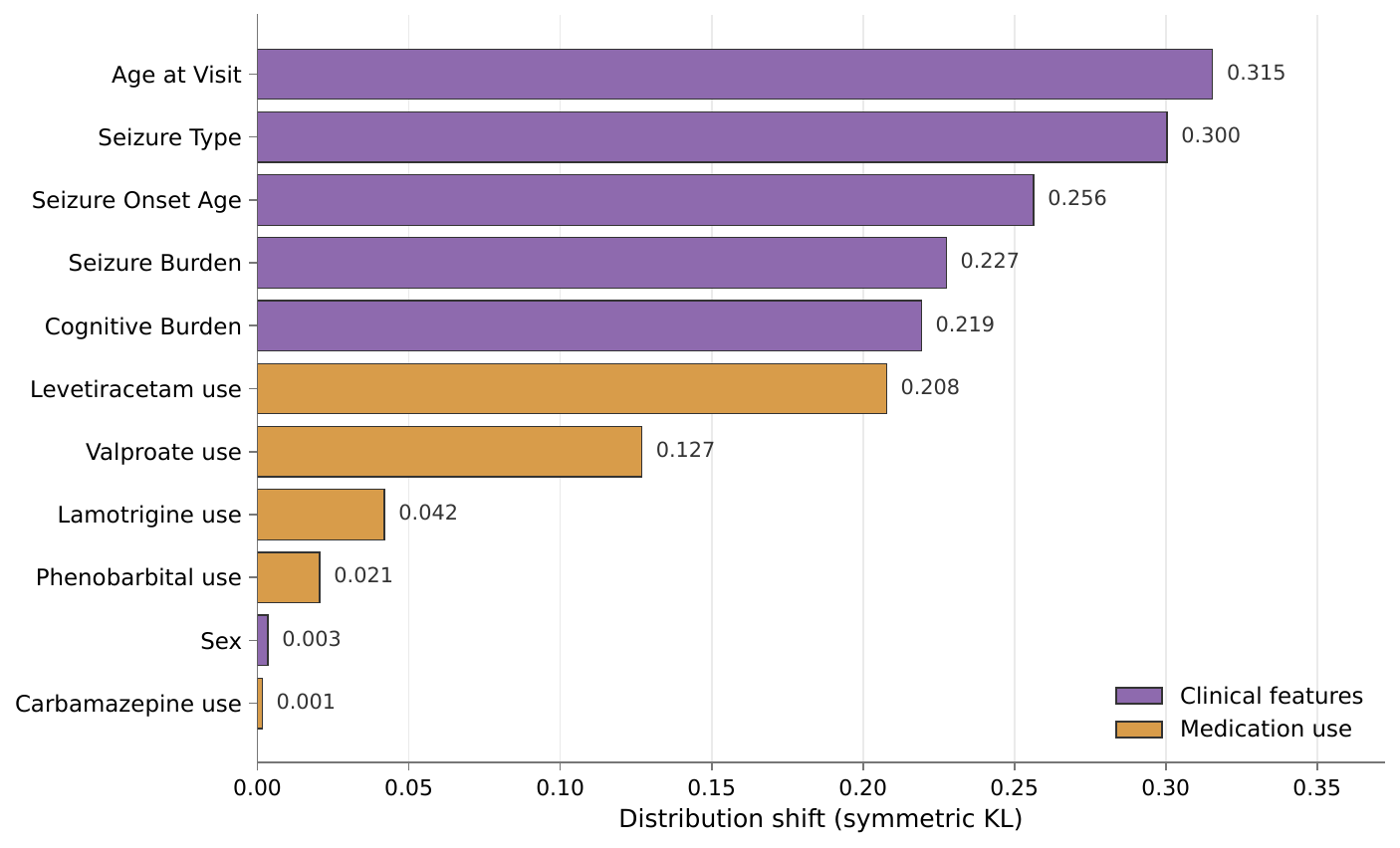}
\caption{Medication-use and documented-feature distribution shift between cohorts. Top: percentage of patients prescribed each anti-seizure medication in Cohort~A and Cohort~B. Bottom: symmetric KL divergence between Cohort~A and Cohort~B for clinical features (purple) and medication-use variables (orange). Larger values indicate larger between-cohort distribution shift.}
\label{fig:cohort_distribution_plots}
\end{figure*}

Table~\ref{tab:cohort_regimen_complexity} summarizes regimen complexity for the main evaluation visits. Visits in the ``None'' row have no extracted active ASM in the physician-prescribed regimen and are excluded from EM@3 and Jaccard calculations. Monotherapy and polytherapy strata are defined by the size of the physician-prescribed active ASM set. Table~\ref{tab:cohort-characteristics} summarizes documented clinical characteristics in the two cohorts. Figure~\ref{fig:cohort_distribution_plots} reports the medication-use fingerprint and documented-feature distribution shift between cohorts.

\begin{table*}[htbp]
\centering
\caption{Clinical characteristics of the two Ugandan epilepsy cohorts. Entries are percentages computed from extracted cohort-profile features. Seizure-type and seizure-frequency rows use documented nonunknown values as denominators; missingness rows report the percentage of visits without the corresponding documented feature.}
\label{tab:cohort-characteristics}
\small
\renewcommand{\arraystretch}{1.08}
\begin{tabular}{@{}l@{\hspace{1.5em}}r@{\hspace{1.4em}}r@{}}
\toprule
Characteristic & Cohort A (\%) & Cohort B (\%) \\
\midrule
\multicolumn{3}{l}{\textit{Demographics}} \\
\quad Male & 61.1 & 65.1 \\
\quad Female & 38.9 & 34.9 \\
\midrule
\multicolumn{3}{l}{\textit{Age at Visit}} \\
\quad Age 0--4 years & 58.4 & 41.4 \\
\quad Age 5--9 years & 27.4 & 28.3 \\
\quad Age 10--17 years & 13.9 & 24.5 \\
\quad Age $\geq$18 years & 0.3 & 5.7 \\
\midrule
\multicolumn{3}{l}{\textit{Seizure Onset}} \\
\quad Onset $<$1 year & 42.2 & 13.6 \\
\quad Onset 1--4 years & 34.5 & 64.4 \\
\quad Onset 5--9 years & 18.5 & 16.9 \\
\quad Onset $\geq$10 years & 4.9 & 5.1 \\
\quad Missing onset age & 13.6 & 83.9 \\
\midrule
\multicolumn{3}{l}{\textit{Epilepsy Type}} \\
\quad Focal seizure type & 36.8 & 72.5 \\
\quad Non-convulsive generalized seizure type & 4.4 & 5.4 \\
\quad Convulsive tonic-clonic seizure type & 58.8 & 22.2 \\
\quad Missing seizure type & 10.8 & 13.9 \\
\midrule
\multicolumn{3}{l}{\textit{Seizure Burden}} \\
\quad Seizure-free at visit & 5.1 & 14.5 \\
\quad Rare seizures ($<$1/month) & 18.3 & 36.2 \\
\quad Monthly seizures & 18.9 & 15.9 \\
\quad Weekly seizures & 14.9 & 15.9 \\
\quad Daily seizures & 42.9 & 17.4 \\
\quad Missing seizure frequency & 47.3 & 81.2 \\
\midrule
\multicolumn{3}{l}{\textit{Cognitive and Developmental Features}} \\
\quad No cognitive/developmental impairment & 44.9 & 68.1 \\
\quad Mild/moderate impairment & 23.5 & 24.3 \\
\quad Severe impairment & 31.6 & 7.6 \\
\bottomrule
\end{tabular}
\end{table*}

Together, Table~\ref{tab:cohort_regimen_complexity} and Figure~\ref{fig:cohort_distribution_plots} show that the two cohorts are related but shifted clinical environments. Cohort~B has a heavier polytherapy profile and a different medication-use fingerprint, whereas Cohort~A is more concentrated in carbamazepine- and valproate-centered prescribing. We therefore report cohort-specific results where relevant rather than treating the combined dataset as homogeneous.

\section{Baseline Definitions}
\label{app:main_results}

This appendix first defines the comparator methods used in Table~\ref{tab:learning_main}. The baselines are ordered by what they test: non-agentic clinical prediction, direct LLM prompting, and frozen-model prompt optimization. This sequence establishes how far simpler approaches go before the specialist audit and expert-derived reference system described in Appendix~\ref{app:consilium_reference}. LLM prompts used for note cleaning, preprocessing, drug-label extraction, and baseline feature extraction are provided with the released code documentation.

\paragraph{EpiPick.}
EpiPick is an external web-based antiseizure-medication selection tool for seizure classification and medication monotherapy~\citep{asadi2020pragmatic,epipick}. The app is designed for patients whose seizures begin at age 10 years or older; since no pediatric regimen-selection baseline exists for this setting, we use it as the closest epilepsy-specific rule-based comparator. To run EpiPick on our visits, we reimplemented the public client-side rule bundle locally, then used an LLM extractor to produce the exact fields required by the algorithm: seizure type, age, gender, menopausal status, and documented modifiers for daily medication use, contraception, tumor, hepatic or renal failure, obesity, diabetes, bleeding risk, neutropenia, renal stones, drug allergy, depression, aggressive behavior, and migraine. The local rule engine maps these extracted variables to EpiPick's ranked medication groups. We evaluate physician-prescribed monotherapy visits only, restrict scoring to the 10 tracked drugs, and count a hit when the prescribed drug appears in EpiPick's Group~1 recommendations.

\paragraph{Naive Bayes.}
The Naive Bayes baseline is a trigram frequency-table model over structured visit features and prescribed medication sets. Before fitting the table, an LLM converts each visit into JSON fields for age, gender, seizure-onset age, seizure frequency, cognitive priority, seizure type, and current-medication flags for the 10 tracked anti-seizure medications. The trigram table conditions on seizure type and seizure frequency. For each training-cell, we estimate a smoothed Bernoulli probability for each drug using $(c+0.5)/(n+1.0)$, where $c$ is the number of visits in that cell whose physician-prescribed regimen includes the drug and $n$ is the number of training visits in the cell. At prediction time, the model uses the most specific cell with at least five training visits, backing off from seizure type plus seizure frequency to seizure type alone, then to the unigram table, and finally to marginal drug probabilities. Let $\mathcal{D}$ be the 10 tracked drugs and let $p_d$ be the selected table's probability for drug $d$. For every candidate regimen $S \subseteq \mathcal{D}$, the model assigns
\[
\mathrm{score}(S)=\sum_{d\in S}\log p_d+\sum_{d\in \mathcal{D}\setminus S}\log(1-p_d).
\]
It enumerates all $2^{10}$ candidate regimens and returns the three highest-scoring sets.

\paragraph{XGBoost clinical.}
The XGBoost clinical baseline tests whether a supervised tabular model can use the same structured visit representation without free-text reasoning. We first run the LLM feature extractor on each visit and retain six clinical variables: age, gender, seizure-onset age, seizure frequency, cognitive priority, and seizure type. Missing values are imputed by the feature mean. On the same patient-level Cohort~A training split used by the learning baselines, we train 10 independent binary XGBoost classifiers, one per tracked drug. For drug $d$, the label is $y_d=1$ if the physician-prescribed regimen includes $d$ and $0$ otherwise. The implementation uses fixed parameters: 200 trees, maximum depth 3, learning rate 0.05, and 0.8 row and column subsampling. At test time, the classifiers produce probabilities $p_d=P(y_d=1\mid x)$ for all 10 drugs. We then score every candidate regimen $S\subseteq\mathcal{D}$ with the same independent-Bernoulli objective used for Naive Bayes,
\[
\mathrm{score}(S)=\sum_{d\in S}\log p_d+\sum_{d\in \mathcal{D}\setminus S}\log(1-p_d),
\]
and return the three highest-scoring regimens.

\paragraph{Base Prompt (0-learning).}
The Base Prompt is the unadapted predictor prompt used to initialize \learnmethod{} and the prompt-optimization baselines; the full template is provided with the released code. This row evaluates that initialization with the learning slot empty: no Inspector feedback, Architect updates, learned rules, specialist analyses, or optimized instructions are supplied. Each visit uses the same longitudinal patient-history input format as the adapted prompt methods. All prompt-based methods are parsed with the same regimen parser; EM@3 counts the visit as correct if any of the three returned drug sets exactly matches the physician-prescribed set. This row isolates what the shared task scaffold achieves before adaptation.

\paragraph{Single-agent.}
The Single-agent baseline is the hand-written direct-prompting comparator for this task; the full prompt is provided with the released code. It was written and refined with clinician input, uses the same longitudinal patient-history input and 10-drug regimen output space as the other prompt-based methods, and performs no learning or optimization.

\paragraph{DSPy.}
DSPy is a prompt-compilation baseline. We instantiate the same clinical task as a DSPy signature with one input field for the longitudinal visit history and three regimen-output fields over the 10 tracked drugs, then optimize the DSPy program with the GEPA optimizer~\citep{agrawal2025gepa}. The optimizer uses the same Cohort~A training and validation split and the same top-3 exact-match objective used for the other prompt-based methods. DSPy therefore tests whether framework-level prompt compilation can improve the shared task scaffold without an explicit clinical memory, Inspector--Architect loop, or cross-patient evidence buffer.

\paragraph{ExpeL.}
ExpeL is an experience-learning agent baseline run through the official ExpeL framework with a \consilium{} task environment. It uses the same 50-patient Cohort~A training split, 20-patient validation split, 10-drug action space, and held-out evaluation protocol as \learnmethod{}. Each episode asks the agent to reason over a visit and finish with three ranked regimens. ExpeL then converts prior episodes into natural-language insights that condition later predictions. We report mean $\pm$ standard deviation across repeated official-code runs.

\paragraph{TextGrad.}
TextGrad is a textual-gradient prompt-optimization baseline initialized from the same 0-learning predictor scaffold. The clinical role, visit input, 10-drug output space, and regimen format are fixed; only the learned-instruction slot is optimized. For each batch, TextGrad runs the predictor, asks an LLM error-diagnosis prompt to compare the prediction with the physician regimen, and applies textual-gradient descent to rewrite the instruction variable. We use the same Cohort~A 50/20 split, batch size 10, 15 update rounds, validation gating, and five-seed reporting protocol as \learnmethod{}.

\section{Interpreting the Main-Results Evaluation}
\label{app:main_results_interpretation}

This appendix reports three interpretive checks on the Table~\ref{tab:learning_main} headline numbers. We first quantify uncertainty in the key paired comparisons, then ask whether longitudinal EM@3 gains reflect more than treatment continuation, and finally examine physician-regimen agreement on visits with documented seizure improvement.

\subsection{Paired confidence intervals for aggregate EM@3}
\label{app:paired_ci}

The standard deviations in Table~\ref{tab:learning_main} describe variation across training seeds, but they do not quantify uncertainty in a difference between methods evaluated on the same patients. We therefore use a paired hierarchical bootstrap with 10{,}000 replicates. Each replicate resamples the five training seeds and test patients with replacement, retaining all visits from each sampled patient and evaluating both methods on the same resampled cases. We report percentile 95\% confidence intervals for the aggregate EM@3 difference; positive values favor \learnmethod{}-Multi.

\begin{table}[H]
\centering
\small
\caption{Paired hierarchical-bootstrap differences in aggregate EM@3. Entries are percentage-point differences with percentile 95\% confidence intervals; positive values favor \learnmethod{}-Multi.}
\label{tab:paired_em3_ci}
\begin{tabular}{lcc}
\toprule
Comparison & Cohort A & Cohort B \\
\midrule
Multi $-$ Base Prompt & \textbf{+38.8} [33.6, 44.0] & \textbf{+33.7} [29.0, 38.4] \\
Multi $-$ Single & \textbf{+4.0} [1.6, 6.4] & +1.2 [$-$1.6, 4.0] \\
Multi $-$ DSPy (GEPA) & \textbf{+3.3} [0.5, 6.0] & +2.1 [$-$0.3, 4.2] \\
\bottomrule
\end{tabular}
\end{table}

\subsection{Later-visit treatment changes}
\label{app:copy_check}

Later visits frequently preserve the active medication set: the physician-prescribed regimen is unchanged from the prior visit in 85.1\% of Cohort~A V2/V3 visits and 76.6\% of Cohort~B V2/V3 visits. Aggregate later-visit agreement therefore mixes continuation and revision decisions. We report each method both on all V2/V3 visits and on the \emph{change-visit} subset, defined as visits where the physician-prescribed active drug set differs from the previous visit.

\begin{table}[!htbp]
\centering
\small
\caption{Later-visit and change-visit performance. Change visits are V2/V3 cases where the physician-prescribed active drug set differs from the previous visit.}
\label{tab:longitudinal}
\begin{tabular}{llcc}
\toprule
Cohort & Method & V2/V3 all EM@3 & Change-only EM@3 \\
\midrule
\multirow{6}{*}{A} & Single-agent & 76.2 & 33.9 \\
& DSPy & \textbf{83.4} & 37.1 \\
& ExpeL & 72.8 & 25.8 \\
& TextGrad & 74.4 & 38.7 \\
& \learnmethod{}-Single & 76.7 & 34.2 \\
& \learnmethod{}-Multi & 81.7 & \textbf{39.0} \\
\midrule
\multirow{6}{*}{B} & Single-agent & 73.2 & 46.4 \\
& DSPy & 79.3 & 50.8 \\
& ExpeL & 77.2 & 51.6 \\
& TextGrad & 69.4 & 41.7 \\
& \learnmethod{}-Single & 74.2 & 47.3 \\
& \learnmethod{}-Multi & \textbf{83.5} & \textbf{55.6} \\
\bottomrule
\end{tabular}
\end{table}

Table~\ref{tab:longitudinal} shows that \learnmethod{}-Multi recovers 39.0\% and 55.6\% of regimens on change visits in Cohorts~A and B, respectively. Methods that look competitive on aggregate later-visit EM@3 have lower change-visit accuracy, especially on Cohort~B (41.7\% for TextGrad and 47.3\% for \learnmethod{}-Single versus 55.6\% for \learnmethod{}-Multi). This pattern indicates that \learnmethod{}-Multi captures a non-trivial revision component rather than only the high-continuity structure of later visits.

\subsection{EM@3 on visits with documented seizure improvement}
\label{app:gt_validity}

Physician-agreement metrics measure concordance with the prescribed regimen. We examine whether this agreement remains strong on visits with documented seizure improvement. The seizure status recorded at a visit describes the preceding treatment period; it does not establish the effectiveness of the prescription selected at that visit. This analysis therefore describes agreement within an improvement-stratified subset, rather than validating the selected regimen's clinical effectiveness.

We use the seizure-frequency annotations available in Cohort~A (`Seizure frequency' for V2 and for V3) to identify visits with documented seizure improvement. We define \emph{Improved} as V2/V3 visits annotated `Reduced' or `Seizure-free' --- visits with documented improvement during the preceding treatment period. This subset contains 342 visits, or 92\% of the V2/V3 visits in Cohort~A for which a seizure-frequency annotation is recorded. Visits annotated `Unchanged' or `Other' are excluded from this improvement-defined subset. Drug-resistant epilepsy --- patients who fail to achieve seizure freedom after adequate trials of two tolerated, appropriately chosen and used antiepileptic drug schedules~\citep{kwan2010definition} --- is well documented in pediatric epilepsy populations~\citep{kumar2021drechildren}, and a proper evaluation on these patients requires longer follow-up, multi-trial outcome data, and likely a reformulated label.

\begin{table}[t]
\centering
\small
\caption{EM@3 on Cohort~A V2/V3 visits with documented seizure improvement (`Improved' = `Reduced' or `Seizure-free' on the per-visit seizure-frequency annotation).}
\label{tab:seizure_freq_strat}
\begin{tabular}{lc}
\toprule
Method & Improved EM@3\\
\midrule
Single-agent & 80.7 \\
DSPy & 84.2 \\
ExpeL & 78.5 \\
TextGrad & 79.7 \\
\learnmethod{}-Single & 82.3 \\
\learnmethod{}-Multi & \textbf{86.0} \\
\bottomrule
\end{tabular}
\end{table}

Table~\ref{tab:seizure_freq_strat} shows that, on this subset with documented seizure improvement, \learnmethod{}-Multi achieves 86.0\% EM@3, outperforming the frozen single-agent baseline (80.7\%) and prompt-learning comparators including TextGrad (79.7\%) and \learnmethod{}-Single (82.3\%). \learnmethod{}-Multi has the highest physician-regimen agreement among the evaluated methods on this subset. These results do not establish clinical effectiveness or a corresponding ranking on visits where seizures worsened.

\section{Single-Agent Audit and Expert System \consilium{}}
\label{app:consilium_reference}

This appendix explains the bridge from direct prompting to the expert-designed \consilium{} reference system. Section~\ref{sec:miscalibration} describes the neurologist audit of single-agent failures, and Section~\ref{sec:consilium} describes the resulting \consilium{} comparator. The single-agent baseline reached non-trivial accuracy, but neurologist review showed that its errors were not random: they recurred along specific clinical lenses. We used that audit to build \consilium{} as the expert-designed comparator that tests whether expert decomposition can correct those errors.

\subsection{Single-Agent Failure Audit}
\label{sec:miscalibration}

The single-agent LLM was a useful starting point because it often produced plausible epilepsy reasoning, but its errors exposed a pattern that direct prompting did not address. Neurologist review showed that the misses clustered around local and longitudinal details: active-medication timelines, seizure classification, pediatric context, drug access, escalation to polytherapy, interaction risk, and infectious triggers. \consilium{} follows from that review: each recurring failure mode becomes a specialist lens, and the final prescription is synthesized across those lenses rather than produced by one monolithic prompt.

\paragraph{Case sampling.}
To understand why standard prompting failed, we intentionally enriched the neurologist audit set for model errors. The goal was not to estimate population-level accuracy, which is reported separately in Table~\ref{tab:learning_main}, but to identify whether the errors reflected recurring, patchable reasoning failures rather than irreducible clinical ambiguity or missing medical knowledge. We sampled 20 longitudinal Cohort~A patients, covering 60 visits total. The sampling criteria were not shown to the neurologists.

\begin{itemize}
    \item 10 polytherapy patients whose regimens were predicted incorrectly.
    \item 5 polytherapy patients whose regimens were predicted correctly.
    \item 3 monotherapy patients whose regimens were predicted incorrectly.
    \item 2 monotherapy patients whose regimens were predicted correctly.
\end{itemize}

\begin{figure}[!htbp]
\centering
\includegraphics[width=0.95\textwidth]{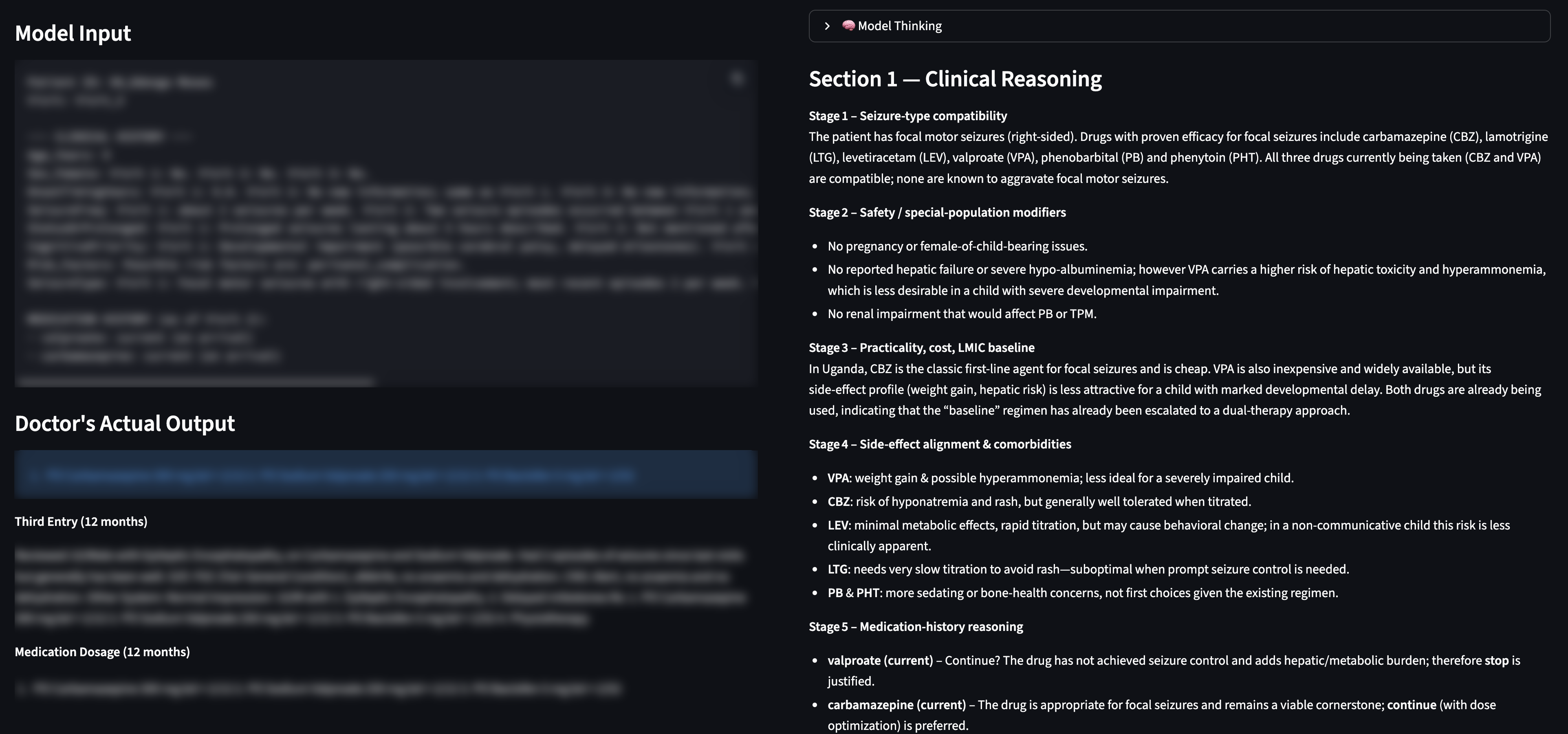}

\vspace{0.35em}

\includegraphics[width=0.95\textwidth]{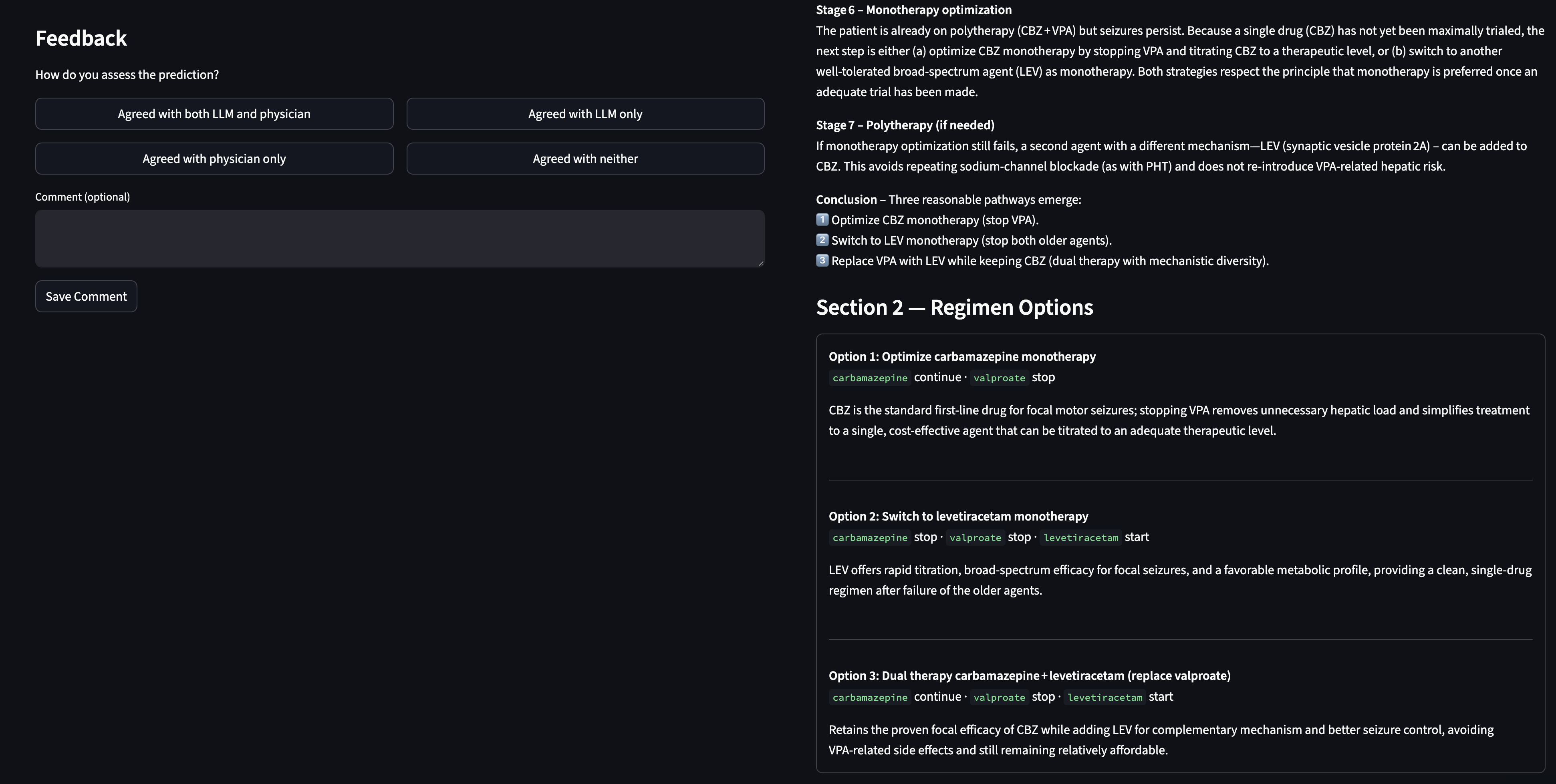}
\caption{Neurologist audit interface. The paper figure is partially redacted: patient identifiers, Model Input text, and Doctor's Actual Output text are blurred. In the review interface, neurologists saw the Model Input, Doctor's Actual Output, Model Thinking, and Regimen Options panels, then selected one of four Feedback choices and could add an optional written comment.}
\label{fig:neurologist_audit_interface}
\end{figure}

\paragraph{Neurologist audit protocol and descriptive results.}
Two pediatric neurologists independently reviewed the sampled cases using the audit interface shown in Figure~\ref{fig:neurologist_audit_interface}. For each visit, the interface showed the \emph{Model Input}, \emph{Doctor's Actual Output}, \emph{Model Thinking}, and \emph{Regimen Options} panels. The neurologists then selected one of four \emph{Feedback} choices---``Agreed with both LLM and physician,'' ``Agreed with LLM only,'' ``Agreed with physician only,'' or ``Agreed with neither''---with an optional written comment. For inter-rater agreement, we used the common binary endpoint available for both neurologists: whether the LLM was accepted in the Feedback choice.

Table~\ref{tab:neurologist_comments} shows representative comments from these reviews. These comments illustrate why aggregate exact-match performance was not enough: the neurologists often agreed that a prediction was clinically plausible while still identifying a recurring reasoning gap, such as timeline drift, missed seizure classification, pediatric dosing, local formulary constraints, or unsafe combinations.

\begin{table}[H]
\centering
\caption{Representative neurologist audit comments. Comments are reproduced verbatim from the audit records; case identifiers are omitted. The final two rows correspond to the redacted case shown in Figure~\ref{fig:neurologist_audit_interface}.}
\label{tab:neurologist_comments}
\footnotesize
\setlength{\tabcolsep}{4pt}
\renewcommand{\arraystretch}{1.12}
\begin{tabularx}{\textwidth}{@{}p{0.145\linewidth}p{0.12\linewidth}Y@{}}
\toprule
Lens suggested & Reviewer & Written comment \\
\midrule
Safety review & Neurologist 1 & Option 2, polytherapy is more appropriate. Reasoning of the model is appropriate and ideal in the context. CBZ, VPA combination is more likely to cause side effects. \\
\addlinespace[0.35em]
Diagnostics & Neurologist 2 & LLM keeps harping on GTCs needing broad spectrum, however focal epilepsy can manifest as GTCs or convert into GTCs from focal features, what the LLM should be picking up is focal versus generalized, rather than the semiology, since in this case the EEG showed focal epilepsy \\
\addlinespace[0.35em]
Local formulary & Neurologist 1 & Agree with LLM reasoning, but also appreciate the neurologist's plan to continue drugs that is working.  VPA is more commonly available than LEV and provided by the national formulary,  and as such more frequently prescribed. One option could be adding a prompt-- where drug availability (which drugs are available in the clinic or affordable by the patient)  input is provided by the provider before the visit. \\
\addlinespace[0.35em]
Pediatric dosing & Neurologist 2 & agree with option three only, LLM should detect that CBZ monotherapy optimized for weight isn't working, agree could consider replacing keppra with depakote.  Since no dose change between visit 2-3 of standing medications I would have double checked that both meds are at the weight based max dosing since children tend to gain weight every 6 months.  Slightly disagree with physician since dosages weren't changed despite having seizures still, but since patient has DRE decision might have been made that increased or optimized dose is futile which is statistically likely. \\
\bottomrule
\end{tabularx}
\end{table}

For the binary endpoint, ``Agreed with both LLM and physician'' and ``Agreed with LLM only'' were coded as LLM accepted; ``Agreed with physician only'' and ``Agreed with neither'' were coded as LLM not accepted. Binary LLM acceptability was summarized across 120 neurologist-visit judgments:

\begin{center}
\small
\begin{tabular}{lcc}
\toprule
Binary Feedback coding & Count & Percentage \\
\midrule
LLM accepted & 104/120 & 86.7\% \\
LLM not accepted & 16/120 & 13.3\% \\
\bottomrule
\end{tabular}
\end{center}

By neurologist, the LLM was accepted at similar rates:

\begin{center}
\small
\begin{tabular}{lcc}
\toprule
Neurologist & Count & Percentage \\
\midrule
Neurologist 1 & 51/60 & 85.0\% \\
Neurologist 2 & 53/60 & 88.3\% \\
\bottomrule
\end{tabular}
\end{center}

Inter-rater agreement on binary LLM acceptability across the 60 shared visits was:

\begin{center}
\small
\begin{tabular}{lc}
\toprule
Agreement measure & Value \\
\midrule
Exact binary agreement & 54/60 (90.0\%) \\
Cohen's $\kappa$ & 0.57 \\
Gwet's AC1 & 0.87 \\
PABAK (prevalence- and bias-adjusted $\kappa$) & 0.80 \\
Positive agreement & 94.2\% \\
Negative agreement & 62.5\% \\
\bottomrule
\end{tabular}
\end{center}

Because the LLM was accepted on most sampled visits, Cohen's $\kappa$ is conservative under class imbalance. Gwet's AC1 and PABAK provide prevalence-robust agreement summaries and show high agreement on the shared binary endpoint. We therefore use the Feedback choices and written comments descriptively, primarily to identify recurrent failure modes rather than to claim clinical superiority. We then characterized the written comments into recurring failure modes, summarized in Table~\ref{tab:failure_modes}.

\begin{table}[H]
\centering
\caption{Recurring failure modes from the neurologist audit. Rows group single-agent reasoning failures from 20 Cohort~A patients (60 visits) and the specialist lens later encoded in \consilium{}.}
\label{tab:failure_modes}
\small
\begin{tabular}{p{0.28\linewidth}p{0.39\linewidth}p{0.23\linewidth}}
\toprule
Failure mode & Single-agent pattern & Missing lens \\
\midrule
Timeline drift & Treats old medications as active or misses between-visit changes & Treatment tracking \\
Seizure classification errors & Follows surface seizure descriptions and misses focal/generalized distinctions & Diagnostics \\
Age and weight drift & Carries forward outdated pediatric context and misses dose adequacy & Pediatric dosing \\
Drug availability blindness & Recommends drugs with limited local access and underuses formulary drugs & Local formulary \\
Monotherapy inertia & Simplifies or avoids escalation when physician trajectory supports polytherapy & Escalation logic \\
Unsafe combinations & Misses sedation and interaction risks in multi-drug regimens & Safety review \\
Infection as seizure trigger & Treats all breakthrough seizures as ASM failure & Tropical differential \\
\bottomrule
\end{tabular}
\end{table}

The audit pointed to a design hypothesis: the model did not need only more generic epilepsy knowledge; it needed more reliable separation of clinical lenses. \consilium{} tests that hypothesis by assigning recurring reasoning gaps to specialist agents, asking for independent reports, and requiring a final synthesis step.

\subsection{Expert-Designed \consilium{} Comparator}
\label{sec:consilium}

\begin{figure}[H]
\centering
\includegraphics[width=\linewidth]{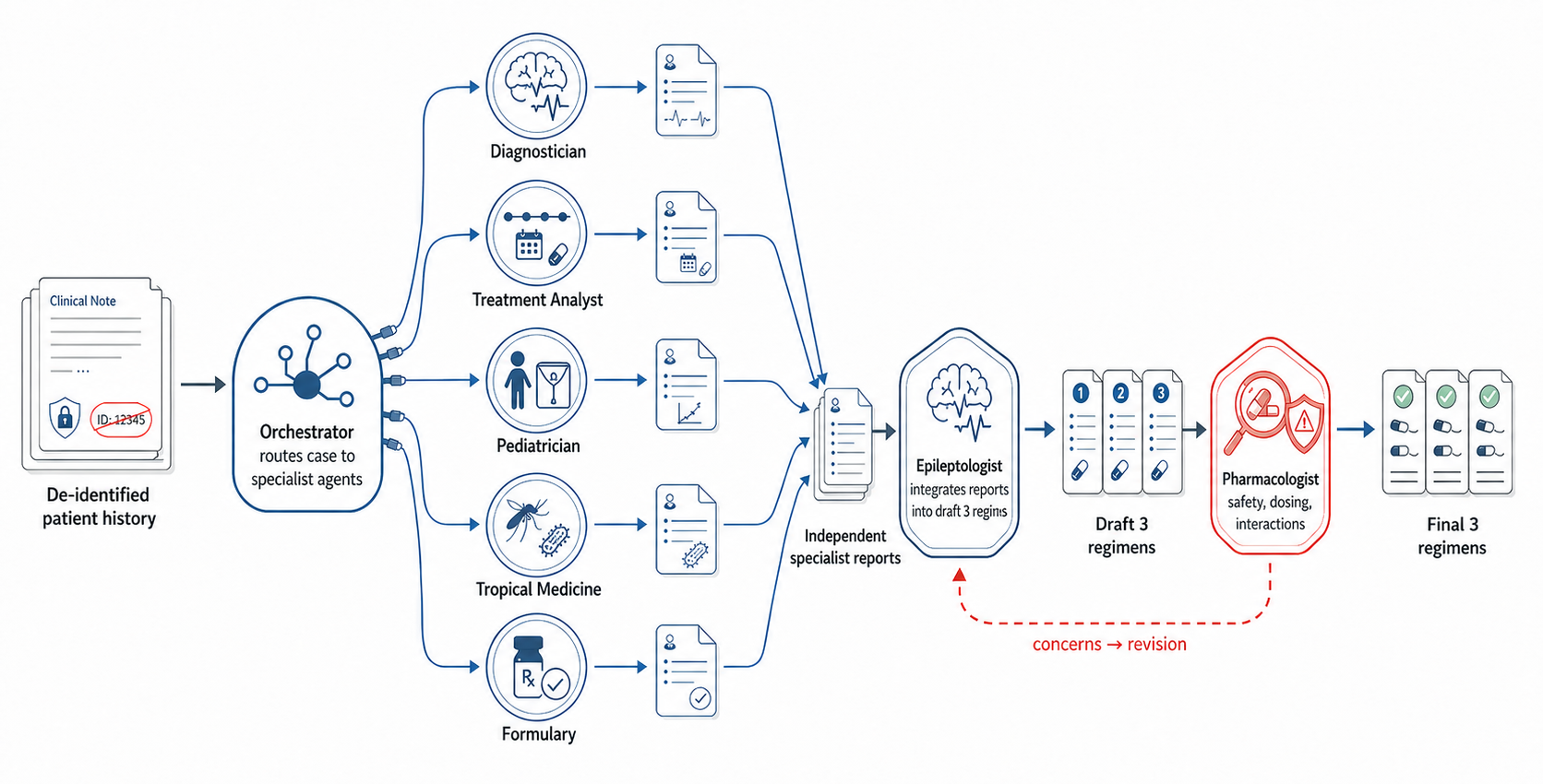}
\caption{\consilium{} expert-system workflow. The orchestrator sends the same visit information to specialist clinical agents, the epileptologist synthesizes their independent reports into ranked regimens, and the pharmacologist reviews safety before final output.}
\label{fig:consilium}
\end{figure}

\consilium{} uses the same visit input as the single-agent baseline, but separates the reasoning before synthesis (Figure~\ref{fig:consilium}). An orchestrator routes the visit to specialist agents, each specialist writes an independent report, and an epileptologist combines those reports into three ranked regimens. A pharmacologist then checks dosing, safety, and interaction risks; if it raises concerns, the critique is returned to the epileptologist before the final regimens are returned. This hand-built system serves as a clinical reference point for \learnmethod{}, not as the scalable method proposed in the main paper.

\paragraph{Aggregate comparison.}
Table~\ref{tab:end_main_em3} focuses on the methods needed to interpret the move from direct prompting to \consilium{}. The Base Prompt row shows the unadapted task scaffold. The \learnmethod{} rows show what can be learned from patient-level supervision without specialist trace feedback. The doctor-designed rows then ask a narrower question: after clinicians helped write a strong single-agent prompt, does explicit specialist decomposition add value beyond that prompt?

\begin{table}[H]
\centering
\small
\setlength{\tabcolsep}{6pt}
\renewcommand{\arraystretch}{1.12}
\caption{Aggregate EM@3 and Jaccard results for prompt-learning and doctor-designed prompt systems. EM@3 is the top-3 exact-match rate; Jaccard is the maximum set overlap between the three predicted regimens and the physician-prescribed regimen. Pre-BPA \learnmethod{} rows are averaged over five seeds; \learnmethod{} + BPA uses the seed-42 trajectory with Beta-Binomial posterior weighting, and its Jaccard is the maximum over the top-3 posterior-weighted regimens.}
\label{tab:end_main_em3}
\begin{tabularx}{\textwidth}{@{}l Y c c c c c c@{}}
\toprule
& & \multicolumn{2}{c}{V1} & \multicolumn{2}{c}{V2} & \multicolumn{2}{c}{V3} \\
\cmidrule(lr){3-4} \cmidrule(lr){5-6} \cmidrule(lr){7-8}
Cohort & Method & EM@3 & Jac & EM@3 & Jac & EM@3 & Jac \\
\midrule
\multirow{9}{*}{A}
  & \multicolumn{7}{l}{\textit{Prompt initialization and learning}} \\
  & Base Prompt (0-learning) & 37.1 & .541 & 50.7 & .635 & 53.1 & .668 \\
  & \learnmethod{}-Single    & 73.8 & .818 & 83.8 & .820 & 84.5 & .848 \\
  & \learnmethod{}-Multi     & 77.1 & .766 & 87.3 & .867 & 89.6 & .874 \\
  & \learnmethod{} + BPA     & 79.0 & .855 & \textbf{91.8} & \textbf{.955} & \textbf{92.6} & \textbf{.963} \\
\addlinespace[0.5em]
  & \multicolumn{7}{l}{\textit{Doctor-designed prompt systems}} \\
  & Single-agent             & 66.0 & .752 & 74.4 & .837 & 75.6 & .855 \\
  & All-agents-combined      & 74.7 & .824 & 81.9 & .907 & 83.4 & .916 \\
  & \consilium{}             & \textbf{79.5} & \textbf{.864} & 88.0 & .945 & 91.0 & .935 \\
\midrule
\multirow{9}{*}{B}
  & \multicolumn{7}{l}{\textit{Prompt initialization and learning}} \\
  & Base Prompt (0-learning) & 44.5 & .592 & 50.7 & .685 & 51.9 & .697 \\
  & \learnmethod{}-Single    & 72.1 & .767 & 83.0 & .849 & 82.4 & .837 \\
  & \learnmethod{}-Multi     & 73.4 & .834 & 83.0 & .911 & 84.3 & .910 \\
  & \learnmethod{} + BPA     & \textbf{76.5} & \textbf{.852} & \textbf{89.8} & \textbf{.952} & \textbf{92.8} & \textbf{.966} \\
\addlinespace[0.5em]
  & \multicolumn{7}{l}{\textit{Doctor-designed prompt systems}} \\
  & Single-agent             & 65.9 & .769 & 70.1 & .826 & 71.8 & .830 \\
  & All-agents-combined      & 64.3 & .742 & 77.7 & .908 & 79.0 & .933 \\
  & \consilium{}             & 71.5 & .818 & 83.0 & .903 & 86.8 & .934 \\
\bottomrule
\end{tabularx}
\end{table}

\noindent\textit{Aggregation.} In Table~\ref{tab:end_main_em3}, aggregate EM@3 is computed over evaluated visit-level cases. Equivalently, it is the visit-count-weighted average of the corresponding monotherapy and polytherapy strata within each cohort and visit. Visits with no extracted active ASM are excluded unless stated otherwise.

\paragraph{All-agents-combined.}
All-agents-combined is the control for prompt content versus council structure. It concatenates the specialist roles used in \consilium{} into one direct prompt, effectively asking a single model call to reason as all specialists at once and return the final ranked regimens in one forward pass. The gap between this row and \consilium{} tests whether the council structure adds value beyond listing the same clinical lenses in a larger prompt.

\paragraph{Ablations test whether the council is reducible.}
Appendix~\ref{app:consilium_ablations} reports leave-one-out and only-one ablations for \consilium{}. The full council is strongest overall, but the leave-one-out rows show that different specialists matter in different cohort--visit slices rather than contributing a uniform gain everywhere.

\section{Cross-Model Transfer Learning}
\label{app:cross_model_transfer}

A learned prompt memory is useful only if it captures reusable task knowledge rather than idiosyncrasies of the model that produced it. We therefore ask whether the lessons learned by a stronger model can improve a smaller model at inference time. In the native 20B condition, the 20B model both learns the prompt memory and uses that memory for evaluation. In the 120B$\to$20B condition, the memory is learned by the 120B model, then inserted unchanged into the 20B inference prompt. The evaluator is always the 20B model; only the source of the learned memory changes. This isolates whether the learned text acts as portable clinical guidance rather than model-specific prompt phrasing.

Table~\ref{tab:transfer} reports the cross-model transfer experiment. Transferred \learnmethod{} memory improves over native 20B learning on five of six cohort--visit pairs. The largest gains occur in Cohort~A, where 120B-derived memory improves all three visits. Cohort~B shows the same pattern at later visits, with a small drop at V1 but gains at V2 and V3. TextGrad also benefits from 120B$\to$20B transfer, but its transferred memories remain less consistently aligned with the clinical task structure than \learnmethod{}.

These results support the view that \learnmethod{} learns an interpretable, reusable correction artifact. The transferred memory acts like portable clinical guidance: it distills recurring prescribing corrections from the source model's adaptation trajectory and makes them available to a smaller model that would otherwise learn weaker memories from the same supervision. This matters for deployment because larger models may be useful for offline adaptation, while smaller models may be preferable for lower-cost or lower-latency inference.

\begin{table}[ht]
\centering
\caption{Cross-model transfer: all variants are evaluated on the 20B model. Native means optimized on 20B; 120B$\to$20B means optimized on 120B and applied to 20B. Entries are mean $\pm$ std over 5 seeds.}
\label{tab:transfer}
\small
\begin{tabular}{clccc}
\toprule
& & \multicolumn{3}{c}{EM@3 (\%)} \\
\cmidrule(lr){3-5}
Cohort & Method & V1 & V2 & V3 \\
\midrule
\multirow{4}{*}{\centering A}
  & TextGrad native              & $61.8_{\pm3.3}$ & $64.5_{\pm5.6}$ & $67.0_{\pm4.7}$ \\
  & TextGrad 120B$\to$20B        & $68.3_{\pm1.7}$ & $71.0_{\pm3.2}$ & $73.0_{\pm1.5}$ \\
  & \learnmethod{} native        & $66.1_{\pm3.2}$ & $69.5_{\pm4.8}$ & $69.9_{\pm2.6}$ \\
  & \learnmethod{} 120B$\to$20B  & $\mathbf{72.5}_{\pm4.2}$ & $\mathbf{71.8}_{\pm5.2}$ & $\mathbf{74.9}_{\pm5.0}$ \\
\midrule
\multirow{4}{*}{\centering B}
  & TextGrad native              & $61.8_{\pm5.1}$ & $68.2_{\pm3.1}$ & $68.5_{\pm2.4}$ \\
  & TextGrad 120B$\to$20B        & $59.9_{\pm3.1}$ & $68.9_{\pm2.9}$ & $69.8_{\pm3.4}$ \\
  & \learnmethod{} native        & $\mathbf{67.8}_{\pm1.6}$ & $72.7_{\pm2.7}$ & $70.6_{\pm4.0}$ \\
  & \learnmethod{} 120B$\to$20B  & $65.8_{\pm3.0}$ & $\mathbf{75.0}_{\pm3.3}$ & $\mathbf{73.5}_{\pm5.6}$ \\
\bottomrule
\end{tabular}
\end{table}

\section{Additional Open Source Model Experiments}
\label{app:llm_backbone_sweep}

The main paper reports results with \texttt{openai.gpt-oss-120b} as the underlying LLM. To check that the \learnmethod{} loop is not specific to a single model family, we replicate the training on additional Bedrock-available LLM backbones spanning a range of scales and architectures. Each backbone is trained with the same single-buffer and multi-agent pipelines, identical 50/20 patient split, and 15 rounds. Test-set inference is run on the round selected by held-out eval accuracy.

\subsection{Models Evaluated}

\begin{itemize}
  \item \texttt{google.gemma-3-12b-it} --- 12B dense, instruction-tuned.
  \item \texttt{mistral.ministral-3-14b-instruct} --- 14B dense.
  \item \texttt{openai.gpt-oss-20b} --- smaller MoE from the same family.
  \item \texttt{google.gemma-3-27b-it} --- 27B dense, instruction-tuned.
  \item \texttt{qwen.qwen3-32b} --- 32B dense.
  \item \texttt{openai.gpt-oss-120b} --- main paper reference (120B, MoE, native reasoning traces).
\end{itemize}

\subsection{Test-Set Performance Across Models}

Table~\ref{tab:llm_backbone_sweep} reports the held-out test performance for each LLM backbone and \learnmethod{} variant.

\begin{table}[ht]
\centering
\caption{EM@3 (\%) of single-buffer and multi-agent \learnmethod{} across LLM backbones, on Cohort A and Cohort B test sets. Best-per-backbone round selected on held-out eval accuracy.}
\label{tab:llm_backbone_sweep}
\small
\begin{tabular}{llcccccc}
\toprule
& & \multicolumn{3}{c}{Cohort A} & \multicolumn{3}{c}{Cohort B} \\
\cmidrule(lr){3-5} \cmidrule(lr){6-8}
Model & Variant & V1 & V2 & V3 & V1 & V2 & V3 \\
\midrule
\multirow{2}{*}{gemma-3-12b}      & Single & 65.3 & 75.7 & 83.2 & 65.9 & 76.5 & 77.8 \\
                                  & Multi  & 59.7 & 79.6 & 81.6 & 59.5 & 71.8 & 75.1 \\
\midrule
\multirow{2}{*}{ministral-3-14b}  & Single & 59.3 & 54.5 & 57.4 & 47.3 & 43.9 & 45.6 \\
                                  & Multi  & 41.1 & 51.8 & 52.0 & 36.3 & 42.3 & 39.6 \\
\midrule
\multirow{2}{*}{gpt-oss-20b}      & Single & 66.1 & 69.5 & 69.9 & 67.8 & 72.7 & 70.6 \\
                                  & Multi  & 65.7 & 72.9 & 77.0 & 70.4 & 75.4 & 78.7 \\
\midrule
\multirow{2}{*}{gemma-3-27b}      & Single & 73.4 & 78.0 & 83.2 & 72.3 & 82.0 & 81.7 \\
                                  & Multi  & 75.7 & 79.3 & 83.5 & 72.3 & 82.8 & 81.0 \\
\midrule
\multirow{2}{*}{qwen3-32b}        & Single & 62.9 & 74.9 & 75.0 & 58.2 & 67.7 & 69.1 \\
                                  & Multi  & 75.8 & 77.6 & 76.6 & 61.6 & 62.2 & 59.5 \\
\midrule
\multirow{2}{*}{gpt-oss-120b}     & Single & 73.8 & 83.8 & 84.5 & 72.1 & \textbf{83.0} & 82.4 \\
                                  & Multi  & \textbf{77.1} & \textbf{87.3} & \textbf{89.6} & \textbf{73.4} & \textbf{83.0} & \textbf{84.3} \\
\bottomrule
\end{tabular}
\end{table}

\subsection{Model Capacity and Specialist Decomposition}

For smaller models such as Gemma-3-12B and Ministral-3-14B, the multi-agent variant often struggled to create useful specialists and write clear, role-specific system prompts. This suggests a capacity-dependent tradeoff: specialist decomposition helps only when the underlying model can reliably manage the extra prompt-construction and coordination steps. Table~\ref{tab:llm_backbone_sweep} reflects this pattern, with Single performing more consistently than Multi on the smallest evaluated models.

\subsection{Medical-Tuned LLM Backbones}
\label{app:medical_backbones}

Because we attribute part of the direct-prompting error to prescribing priors
miscalibrated to the local setting rather than to missing medical knowledge, we
compare against clinical/biomedical LLM backbones under the identical Base Prompt,
output schema, and regimen parser, changing only the model. Table~\ref{tab:medical_backbones}
reports EM@3 stratified by regimen complexity. The medical-tuned backbones improve
substantially over the general \texttt{gpt-oss-120b} Base Prompt, consistent with
their biomedical training, but still fall short of \texttt{gpt-oss-120b} with
\learnmethod{} on both cohorts and both strata. This supports our central claim: the
gain comes from the learning procedure, not from a medically pretrained backbone.

\begin{table}[ht]
\centering
\caption{EM@3 (\%) by backbone under the same Base Prompt, output schema, and parser,
stratified into monotherapy and polytherapy visits across V1--V3. \learnmethod{}
(\texttt{gpt-oss-120b} + Single/Multi) is shown for reference. Best per column in bold.}
\label{tab:medical_backbones}
\small
\begin{tabular}{lcccccc}
\toprule
& \multicolumn{3}{c}{Monotherapy} & \multicolumn{3}{c}{Polytherapy} \\
\cmidrule(lr){2-4} \cmidrule(lr){5-7}
Backbone & V1 & V2 & V3 & V1 & V2 & V3 \\
\midrule
\multicolumn{7}{l}{\textit{Cohort A}} \\
\texttt{gpt-oss-120b}, Base & 44.4 & 56.4 & 59.3 & 10.1 & 29.3 & 30.2 \\
MedGemma-27B                & 67.7 & 80.5 & 84.5 & 22.6 & 46.7 & 54.5 \\
OpenBio-70B                 & 64.6 & 79.0 & 86.6 & 20.8 & 45.0 & 62.3 \\
HuatuoGPT-72B               & 67.7 & 82.6 & 86.9 & 17.0 & 48.3 & 42.9 \\
\texttt{gpt-oss-120b} + \learnmethod{}-Single & 87.0 & 92.1 & 93.2 & 25.3 & 57.0 & 64.1 \\
\texttt{gpt-oss-120b} + \learnmethod{}-Multi & \textbf{89.7} & \textbf{95.3} & \textbf{97.0} & \textbf{30.8} & \textbf{61.7} & \textbf{72.6} \\
\midrule
\multicolumn{7}{l}{\textit{Cohort B}} \\
\texttt{gpt-oss-120b}, Base & 53.1 & 57.6 & 58.4 & 24.8 & 29.1 & 30.4 \\
MedGemma-27B                & 70.3 & 77.6 & 80.3 & 32.3 & 54.8 & 58.2 \\
OpenBio-70B                 & 64.1 & 66.2 & 67.6 & 34.0 & 43.0 & 54.0 \\
HuatuoGPT-72B               & 70.6 & 80.9 & 84.3 & 30.9 & 46.9 & 45.2 \\
\texttt{gpt-oss-120b} + \learnmethod{}-Single & 82.2 & 92.8 & 92.2 & \textbf{48.2} & \textbf{65.2} & 66.4 \\
\texttt{gpt-oss-120b} + \learnmethod{}-Multi & \textbf{87.3} & \textbf{95.1} & \textbf{93.2} & 40.2 & 61.1 & \textbf{69.8} \\
\bottomrule
\end{tabular}
\end{table}

\section{\learnmethod{} Component Ablations}
\label{app:component_ablations}

To isolate which components of the \learnmethod{} loop drive performance, we run paired ablations against the full pipeline. Each ablation removes a single component while keeping the rest of the system fixed. We report no-buffer and no-inspector results on both the single-buffer and multi-agent variants of \learnmethod{}, evaluated on Cohort A and Cohort B with the same train/eval split (50 train, 20 eval) and 15 rounds. We additionally define a no-architect ablation for the single-buffer variant below. Results are reported in Table~\ref{tab:component_ablations}.

\paragraph{No-buffer.} The append-only candidate buffer is removed while the learned prompt state persists across rounds (shared learnings in the single-buffer variant; the agent population in the multi-agent variant). The Architect receives the current prompt state and the current batch's Inspector reports, but no candidate learnings from previous rounds. This isolates the contribution of cross-round candidate memory from the learned state itself.

\paragraph{No-inspector.} The Inspector is removed. The Architect receives raw per-case triples \texttt{(patient notes, predictor's three options, ground-truth prescription)} instead of structured Inspector reports, and must perform error attribution itself. This isolates whether the diagnostic decomposition (match status, error type, root cause, missed signal) is what makes the Architect's updates useful, or whether raw signal suffices.

\paragraph{No-architect.} In the single-buffer variant, the Architect is removed and each Inspector-produced \texttt{CANDIDATE\_LEARNING} is appended directly to the shared rule list. This ablation keeps supervised error diagnosis but removes the consolidation layer: there is no quorum check, deduplication, rewriting, or cross-case synthesis before a lesson enters the Predictor prompt. We do not define an analogous multi-agent no-architect row because the Architect is the mechanism that creates, edits, and prunes specialist agents; removing it from an initially empty multi-agent state would leave no learned agents and collapse the method to the base Predictor rather than a meaningful multi-agent ablation.

\paragraph{No-quorum.} The Inspector, candidate buffer, and Architect are all retained, but the explicit quorum rule is stripped from the Architect's prompt. Concretely, the Architect still consumes structured Inspector reports and the cross-round candidate buffer and still issues prompt-state updates, but it is no longer instructed to require evidence from multiple cases before promoting a candidate learning into the rule list (single-buffer) or before creating, editing, or retiring an agent (multi-agent). This ablation isolates the contribution of the multi-case evidence threshold from the rest of the consolidation layer (deduplication, rewriting, cross-case synthesis), which the Architect continues to perform.

\paragraph{In-context learning.} We also evaluate a non-learning in-context baseline for the single-buffer variant. In this setting, the self-learning loop is disabled entirely: there is no Inspector, Architect, buffer, or learned rule consolidation. Instead, the Predictor receives raw training examples directly in its prompt, where each example consists of a patient record paired with the doctor's ground-truth prescription. This ablation asks whether performance comes simply from exposing the model to many labeled cases at inference time, rather than from distilling those cases into compact, reusable clinical learnings. We report this only for the single-buffer setting because the intervention is directly analogous to replacing the shared learned rule list with labeled in-context examples; there is no corresponding multi-agent state to populate without an Architect.

\begin{table}[ht]
\centering
\caption{Component ablations on the single-buffer and multi-agent variants of \learnmethod{}. EM@3 (\%) on the held-out test sets. The ``Full'' row reproduces the corresponding row of Table~\ref{tab:end_main_em3}.}
\label{tab:component_ablations}
\small
\begin{tabular}{llcccccc}
\toprule
& & \multicolumn{3}{c}{Cohort A} & \multicolumn{3}{c}{Cohort B} \\
\cmidrule(lr){3-5} \cmidrule(lr){6-8}
Variant & Ablation & V1 & V2 & V3 & V1 & V2 & V3 \\
\midrule
\multirow{6}{*}{Single}
  & in-context        & 69.4 & 70.0 & 57.9 & 61.1 & 82.4 & 76.5 \\
  & no-buffer         & 71.8 & 77.6 & 82.0 & 69.8 & 80.7 & 76.0 \\
  & no-inspector      & 71.8 & 80.4 & 81.2 & 65.5 & 80.4 & 80.5 \\
  & no-architect      & 72.6 & 81.6 & 83.2 & 70.6 & 80.1 & 72.4 \\
  & no-quorum         & 71.8 & 79.8 & 82.7 & 61.2 & 72.7 & 78.8 \\
  & Full              & 73.8 & 83.8 & 84.5 & 72.1 & 83.0 & 82.4 \\
\midrule
\multirow{4}{*}{Multi}
  & no-buffer         & 70.6 & 77.6 & 80.9 & 71.1 & 81.2 & 80.8 \\
  & no-inspector      & 66.1 & 76.9 & 84.0 & 72.8 & 81.2 & 80.6 \\
  & no-quorum         & 67.7 & 82.5 & 80.6 & 69.4 & 76.8 & 77.4 \\
  & Full              & 77.1 & 87.3 & 89.6 & 73.4 & 83.0 & 84.3 \\
\bottomrule
\end{tabular}
\end{table}

\section{Clinician Review of \learnmethod{}}
\label{app:manana_clinician_review}

We conducted an additional neurologist audit comparing the expert-designed \consilium{} system with the self-learning \learnmethod{}-Multi system. The reviewer examined 20 longitudinal patients, with three visits per patient and outputs from both systems, yielding $20 \times 3 \times 2 = 120$ system-visit reviews. The review asked whether each output correctly identified seizure type, assessed seizure activity, accounted for medication context, and gave clinically sound drug-selection reasoning; each question was answered as \textit{Yes}, \textit{Partially}, or \textit{No}. The reviewer also rated overall usefulness as \textit{Very useful}, \textit{Somewhat useful}, or \textit{Not useful}.

\textbf{Qualitative findings.}
Because the neurologist's ratings were near ceiling across the answered fields, we use the review primarily as a qualitative audit. We therefore focus on the neurologist's comments, which describe how the systems behave as practical decision-support tools. The clearest theme was option coherence: \learnmethod{}-Multi often proposed related variants of the same clinical plan, while \consilium{} sometimes proposed individually plausible but directionally different options. In Cohort~A, the reviewer wrote that ``System B reasoning is easier to follow,'' and in another case noted that ``all three recommendations are concordant,'' contrasting this with System~A's ``appropriate but different choices.'' The comments make the practical issue clearer: when \consilium{} offered plausible but different options, the reviewer treated this as less useful than recommendations that preserved the same treatment direction.

These observations motivate a deeper system-level audit of when model recommendations are clinically coherent, which we leave as a planned next step.

\section{\learnmethod{} Bayesian Prompt Averaging Ablations}
\label{app:bpa_ablations}

\subsection{Beta-Binomial BPA Derivation}

The main text uses a Beta-Binomial marginal likelihood to score each retained memory state. For a validation case evaluated under memory state $m_k$, there are three relevant outcomes: the physician regimen appears in candidate position 1, appears in candidate positions 2--3, or is absent from the three candidates. Let $c_{k,1}$, $c_{k,>1}$, and $u_k$ denote the corresponding counts, and let $h_k=c_{k,1}+c_{k,>1}$ be the number of top-3 hits.

We factor this outcome into a state-specific top-3 hit probability $\theta_k$ and a conditional candidate-position distribution $\pi$. Thus a case is a top-3 hit with probability $\theta_k$; conditional on a hit, its position is drawn from $\pi$. With $\pi_{>1}=\pi_2+\pi_3$, the validation likelihood is
\[
p(\mathcal{D}_{\mathrm{val}}\mid \theta_k,\pi,m_k)
=
\pi_1^{c_{k,1}}\pi_{>1}^{c_{k,>1}}
\theta_k^{h_k}(1-\theta_k)^{u_k}.
\]
We estimate $\pi=(0.85,0.11,0.04)$ once from a small subset of the training set and treat it as a fixed empirical-Bayes-style plug-in estimate. For the state-specific hit probability, we use the conjugate prior $\theta_k\sim\mathrm{Beta}(1,1)$ and integrate out $\theta_k$:
\[
p(\mathcal{D}_{\mathrm{val}}\mid m_k)
=
\pi_1^{c_{k,1}}\pi_{>1}^{c_{k,>1}}
\int_0^1 \theta_k^{h_k}(1-\theta_k)^{u_k}\,d\theta_k
=
\pi_1^{c_{k,1}}\pi_{>1}^{c_{k,>1}}B(h_k+1,u_k+1).
\]
This is the score used for the Beta-Binomial BPA weights in the main text. The Dirichlet-Multinomial variant below relaxes the shared fixed $\pi$ assumption by estimating a separate candidate-position distribution for each memory state under a conjugate Dirichlet prior.

\subsection{Confidence and deferral curves}
\label{app:bpa_confidence}

The reliability panels below bin cases by BPA posterior confidence and compare each bin's empirical top-1 accuracy against the diagonal of perfect calibration. The deferral curve reports top-1 precision on the non-deferred subset as lower-confidence cases are routed to clinician review.

\begin{figure*}[t]
\centering
\includegraphics[width=\textwidth]{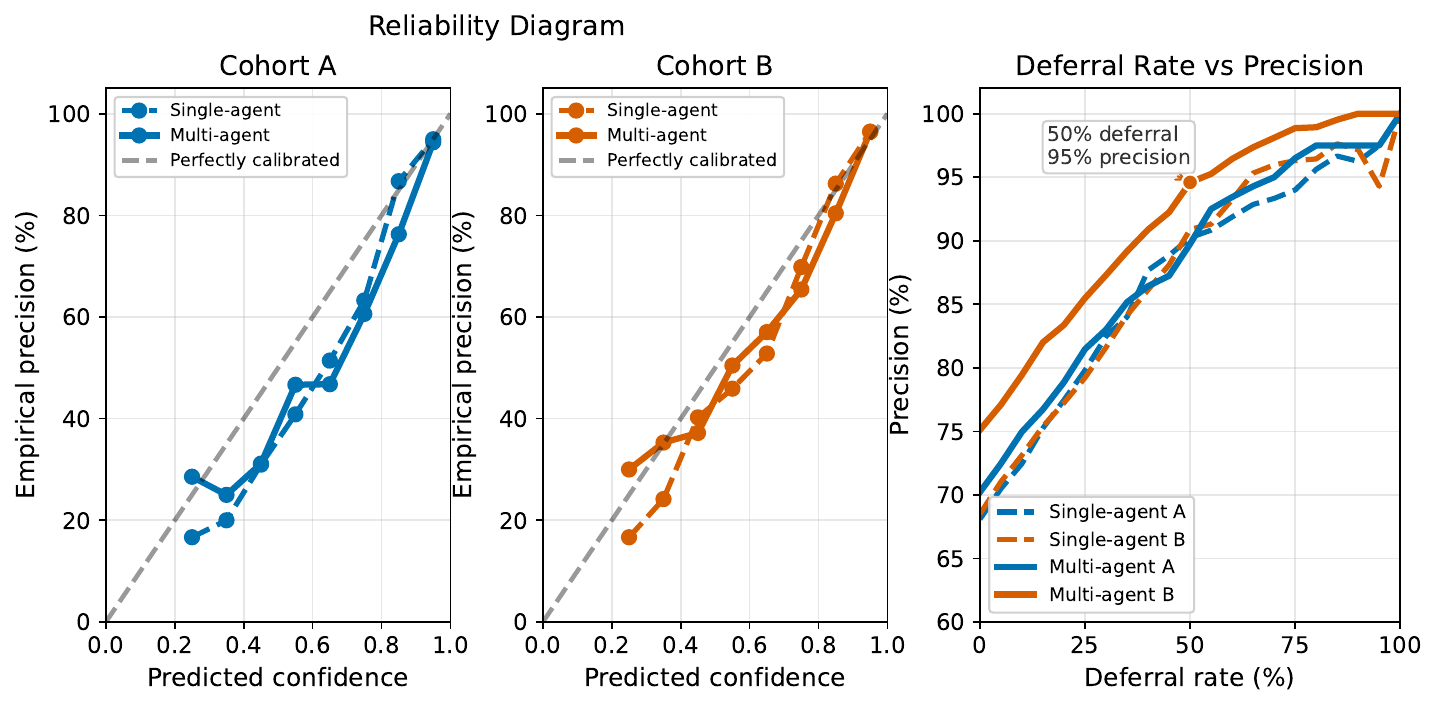}
\caption{BPA confidence and deferral behavior. Dashed and solid curves denote \learnmethod{}-Single and \learnmethod{}-Multi. Left and middle: BPA confidence, defined as the final BPA probability assigned to the top-ranked active ASM regimen, tracks empirical top-1 physician-regimen agreement; the dotted diagonal indicates perfect calibration. Right: deferral curve obtained by deferring lower-confidence cases to clinician review. The marked Cohort~B point shows that at 50\% deferral, \learnmethod{}-Multi reaches 95\% top-1 agreement on the remaining cases.}
\label{fig:bpa_uncertainty}
\end{figure*}

\subsection{Weighting Ablations}

Table~\ref{tab:bpa_ablation} compares the Beta-Binomial BPA used in the main text against two alternatives for the \learnmethod{}-Single trajectory. Linear weighting uses validation top-3 rates directly. Dirichlet-Multinomial BPA relaxes the shared fixed candidate-position prior assumption by estimating a separate candidate-position distribution for each prompt. The three variants produce similar predictions, which suggests that the deferral signal is not an artifact of a fragile weighting rule. Gap is the difference in mean confidence between correct and incorrect top-1 predictions; P@25\% and P@50\% report top-1 precision after retaining the most confident 25\% or 50\% of cases. We use Beta-Binomial BPA in the main text because it gives the clearest closed-form likelihood and the strongest or near-strongest calibration on the full test set.

\begin{table}[ht]
\centering
\small
\setlength{\tabcolsep}{8pt}
\renewcommand{\arraystretch}{1.12}
\caption{BPA weighting ablations for \learnmethod{}-Single on full held-out test sets.}
\label{tab:bpa_ablation}
\begin{tabular}{@{}llccccc@{}}
\toprule
Cohort & Scheme & Top-1 & Top-3 & Gap & P@25\% & P@50\% \\
\midrule
\multirow{3}{*}{A} & Linear            & 76\% & \textbf{88\%} & 0.163 & \textbf{96\%} & 91\% \\
                   & Beta-Binomial BPA & 76\% & \textbf{88\%} & \textbf{0.171} & 95\% & 91\% \\
                   & Dirichlet BPA     & 76\% & 87\% & 0.164 & \textbf{96\%} & 91\% \\
\midrule
\multirow{3}{*}{B} & Linear            & 71\% & \textbf{88\%} & 0.154 & \textbf{95\%} & 85\% \\
                   & Beta-Binomial BPA & \textbf{72\%} & 87\% & 0.150 & \textbf{95\%} & \textbf{86\%} \\
                   & Dirichlet BPA     & 71\% & 87\% & \textbf{0.159} & 94\% & 85\% \\
\bottomrule
\end{tabular}
\end{table}

\section{MIMIC-IV}
\label{app:mimic}

\textbf{Note on task mismatch.}
MIMIC-IV is not the same clinical setting as our primary Ugandan epilepsy cohorts. Our main task uses longitudinal outpatient epilepsy-care notes written before or during serial clinic visits, where the model must infer a locally appropriate anti-seizure medication (ASM) regimen from sparse visit documentation. The MIMIC data are retrospective hospital discharge summaries from a high-resource US hospital system. These notes are written after the admission has ended, so they often contain the treatment course, response to therapy, discharge intent, and medication reconciliation. Therefore, unprocessed MIMIC discharge notes can leak the answer.

The MIMIC cohort is still useful because it is the closest public, credentialed-access source containing epilepsy-related clinical notes and medication records. We use it as a reproducibility and robustness setting, not as a replacement for the Ugandan serial-care task. The cohort is adult and hospital-based: median age is 49 years, IQR 34--60, with 55.3\% female admissions. Across the final MIMIC cohort, patients have broad inpatient medication exposure: median 16 unique prescribed drug strings during the admission, and median 8 active scheduled medications at discharge. The anti-seizure component is only part of that medication burden.

\textbf{Cohort construction.}
We filter MIMIC-IV admissions to those with a primary epilepsy ICD code (345.x or G40.x at \texttt{seq\_num=1}) and Neurology as the first managing service (\texttt{NMED}), keeping the longest discharge summary per admission. After ground-truth (GT) extraction, action-space restriction, and removal of manually audited leaky admissions, the final cohort contains 1{,}977 admissions from 1{,}257 patients (31.6\% monotherapy, 68.4\% polytherapy; max GT regimen size 4). Full filter logic and per-stage counts are in the released code.

\textbf{Ground-truth extraction.}
Ground truth is extracted from the prescription table, not the note: a prescription qualifies as a target ASM if it is oral/enteral, scheduled (not PRN), active at discharge, and newly started during the admission (fosphenytoin excluded). The MIMIC action space is the 15 ASMs meeting the 4\% frequency threshold in this cohort: levetiracetam, lamotrigine, lacosamide, zonisamide, phenytoin, valproate, oxcarbazepine, lorazepam, gabapentin, clobazam, topiramate, clonazepam, carbamazepine, pregabalin, phenobarbital.

\textbf{Cleaning rationale.}
MIMIC discharge summaries are written after the admission and encode the prescribed regimen through hospital-course text, medication reconciliation, and discharge planning. The cleaner removes those sections so the input contains pre-admission and admission-evidence text without the discharge-side decisions.

\textbf{LLM cleaning.}
Cleaning is performed by an LLM following the released editing prompt, with an additional regex pass that strips residual discharge-tail headers (e.g., \emph{Discharge Diagnosis}, \emph{Discharge Medications}, \emph{Follow-up Instructions}) and excludes manually audited leaky admissions. Results are reported on this configuration; the saved test predictions and prompt are released for reproducibility.

\textbf{Held-out test results.}
Table~\ref{tab:mimic_results} reports held-out test performance from the MIMIC-IV test split (split seed 42). Each learning system uses the same 50-patient training split and 20-patient validation split; \learnmethod{}-Single, \learnmethod{}-Multi, TextGrad, and ExpeL are evaluated from their best learned checkpoints. The Base Prompt row evaluates the unadapted single-agent template against MIMIC.

\begin{table}[ht]
\centering
\small
\setlength{\tabcolsep}{6pt}
\renewcommand{\arraystretch}{1.12}
\caption{MIMIC-IV held-out test results. EM@1 and EM@3 are top-1 and top-3 exact-match rates over the 15-ASM MIMIC action space. Mono@3 and Poly@3 stratify EM@3 by ground-truth regimen size. EpiPick is evaluated on all monotherapy admissions, Poly@3 does not apply.}
\label{tab:mimic_results}
\begin{tabular}{@{}l c c c c c@{}}
\toprule
Method & EM@1 & EM@3 & Jac & Mono@3 & Poly@3 \\
\midrule
\multicolumn{6}{l}{\textit{Expert System}} \\
EpiPick                          & 18.4 & 40.5 & .405 & 40.5 & --- \\
\addlinespace[0.5em]
\multicolumn{6}{l}{\textit{Classical methods}} \\
Naive Bayes                    & 9.0 & 16.9 & .360 & 43.6 & 4.5 \\
\addlinespace[0.5em]
\multicolumn{6}{l}{\textit{Direct LLM prompting}} \\
Base Prompt (0-learning)         & 30.7 & 36.0 & .572 & 57.7 & 24.5 \\
\addlinespace[0.5em]
\multicolumn{6}{l}{\textit{Prompt optimization}} \\
ExpeL                            & 28.7 & 28.7 & .480 & 32.7 & 26.5 \\
TextGrad                         & 28.0 & 32.7 & .580 & 48.1 & 24.5 \\
\learnmethod{}-Single       & 31.3 & 40.0 & \textbf{.600} & \textbf{65.4} & 26.5 \\
\learnmethod{}-Multi        & \textbf{34.0} & \textbf{47.3} & .566 & 61.5 & \textbf{39.8} \\
\bottomrule
\end{tabular}
\end{table}

\section{\consilium{} Council Ablations}
\label{app:consilium_ablations}

Table~\ref{tab:ablation} reports leave-one-out and only-one ablations for \consilium{}.

\begin{table}[ht]
\centering
\caption{Ablations of the \consilium{} council. Entries are EM@3 percentages across visits in each cohort.}
\label{tab:ablation}
\small
\begin{tabular}{lcccccc}
\toprule
& \multicolumn{3}{c}{Cohort A} & \multicolumn{3}{c}{Cohort B} \\
\cmidrule(lr){2-4} \cmidrule(lr){5-7}
Configuration & V1 & V2 & V3 & V1 & V2 & V3 \\
\midrule
\consilium{} (full)       & 79.5          & \textbf{88.0} & \textbf{91.0} & 71.5 & \textbf{83.0} & \textbf{86.8} \\
\midrule
w/o Diagnostician      & 78.3          & 86.1          & 90.1          & 66.2          & 72.4          & 79.0 \\
w/o Treatment Analyst  & 75.0          & 85.4          & 83.9          & 67.5          & 73.3          & 81.7 \\
w/o Formulary          & 74.6          & 87.6          & 85.0          & \textbf{73.1} & 73.3          & 82.3 \\
w/o Tropical Medicine  & 75.7          & 86.9          & 89.0          & 66.6          & 71.0          & 82.3 \\
w/o Pediatrician       & \textbf{80.5} & 83.6          & 90.1          & 61.0          & 77.1          & 80.2 \\
\midrule
Only Treatment Analyst & 73.2          & 86.5          & 87.9          & 68.2          & 76.5          & 81.4 \\
Only Diagnostician     & 68.8          & 78.8          & 76.2          & 70.2          & 73.0          & 79.6 \\
Only Formulary         & 76.1          & 79.2          & 79.5          & 63.6          & 73.0          & 76.0 \\
Only Tropical Medicine & 70.6          & 80.7          & 79.9          & 64.6          & 78.9          & 79.9 \\
Only Pediatrician      & 73.5          & 80.3          & 83.5          & 65.9          & 77.7          & 81.1 \\
\midrule
Single-agent           & 66.0          & 74.4          & 75.6          & 65.9          & 70.1          & 71.8 \\
\bottomrule
\end{tabular}
\end{table}

\clearpage
\section{Learned Artifacts Across Optimization Methods}
\label{app:learned_artifacts}

We selected representative artifacts from high-performing runs to illustrate the form of each method's learned text state. The full artifacts are provided with the released code documentation; here we summarize the qualitative differences that matter for interpreting the results.

\textbf{TextGrad: detailed prose, weak grounding.} TextGrad learns by rewriting a global instruction variable. In this setting, the learned artifact can read like a guideline document, but the supervision signal is only a prescribed drug set. In representative runs, it commits to dose ranges, laboratory schedules, formulation choices, pill-burden constraints, stocking assumptions, and cost hierarchy details that are not recoverable from the training labels alone.

\textbf{ExpeL: useful but generic rules.} ExpeL produces a ranked experience memory. Its lessons capture plausible prescribing heuristics, such as prioritizing continuation, selecting locally common first-line agents, and using weight-based dosing, but they remain at the level of generic best practice rather than naming the recurring drug-specific or seizure-specific local patterns that drive cohort-level errors.

\textbf{DSPy-GEPA: a sharper interface, no clinical memory.} DSPy-GEPA produces an optimized instruction for the prediction program rather than a persistent clinical memory. It improves the task interface by emphasizing explicit-directive parsing, output formatting, dose-adequacy reasoning, and constraints against unnecessary drug additions, but it has no slot for cross-patient lessons to accumulate over rounds.

\textbf{\learnmethod{}-Single: compact, evidence-gated rules.} The Single variant learns a short list of global correction rules that survive the Architect's quorum check across cases. The representative rules name specific drugs and clinical situations, such as continuing a tolerated regimen when no modification directive is present, adding levetiracetam before changing tolerated carbamazepine in partially controlled focal epilepsy, and avoiding escalation after a solitary fever-related breakthrough seizure.

\textbf{\learnmethod{}-Multi: specialists, not prescribers.} The Multi variant turns recurring errors into specialist signal extractors. The learned agents surface clinical observations for the Predictor rather than directly writing prescriptions, making the memory easier to audit as clinical signal extraction instead of a hidden decision policy.

\end{document}